\pdfoutput=1
\PassOptionsToPackage{table,xcdraw}{xcolor}
\documentclass[11pt]{article}

\usepackage[preprint]{acl}

\usepackage{times}
\usepackage{latexsym}
\usepackage[T1]{fontenc}
\usepackage[utf8]{inputenc}
\usepackage{microtype}
\usepackage{inconsolata}
\usepackage{graphicx}
\usepackage{booktabs}
\usepackage{xspace}
\usepackage[capitalise]{cleveref}

\newcommand{\Fone}{F\textsubscript{1}\xspace}
\newcommand{\fonesd}[2]{#1\,\textsubscript{$\pm$#2}}

\newcommand{\modelname}{\textsc{ltz-E1}\xspace}
\newcommand{\luxembert}{\textsc{LuxemBERT}\xspace}
\newcommand{\mbert}{\textsc{mBERT}-base\xspace}
\newcommand{\mmbert}{\textsc{mmBERT}-base\xspace}
\newcommand{\roberta}{\textsc{XLM-R}-base\xspace}

\newcommand{\modelnames}{\textsc{ltzE1b}\xspace}
\newcommand{\modelnamesm}{\textsc{ltzE1m}\xspace}
\newcommand{\luxemberts}{\textsc{LxBRT}\xspace}
\newcommand{\mberts}{\textsc{mBRT}\xspace}
\newcommand{\mmberts}{\textsc{mmBRT}\xspace}
\newcommand{\robertas}{\textsc{XLMR}\xspace}

\newcommand{\ner}{\textsc{NER}\xspace}
\newcommand{\tc}{\textsc{TC}\xspace}
\newcommand{\hc}{\textsc{HA}\xspace}
\newcommand{\lam}{\textsc{LA (multi)}\xspace}
\newcommand{\lab}{\textsc{LA (binary)}\xspace}
\newcommand{\sid}{\textsc{ID}\xspace}
\newcommand{\sa}{\textsc{SA}\xspace}
\newcommand{\rte}{\textsc{RTE}\xspace}

\newcommand{\lams}{\textsc{LAM}\xspace}
\newcommand{\labs}{\textsc{LAB}\xspace}

\title{\textsc{ltzGLUE}: Luxembourgish General Language Understanding Evaluation}

\author{
 \textbf{Alistair Plum\textsuperscript{1}},
 \textbf{Felicia Körner\textsuperscript{2,3}},
 \textbf{Anne-Marie Lutgen\textsuperscript{1}},
 \textbf{Laura Bernardy\textsuperscript{1}},
\\
 \textbf{Fred Philippy\textsuperscript{1}},
 \textbf{Emilia Milano\textsuperscript{1}},
 \textbf{Nils Rehlinger\textsuperscript{1}},
 \textbf{Cédric Lothritz\textsuperscript{4}},
\\
 \textbf{Tharindu Ranasinghe\textsuperscript{5}},
 \textbf{Barbara Plank\textsuperscript{2,3}},
 \textbf{Christoph Purschke\textsuperscript{1}}
\\
\\
 \textsuperscript{1}University of Luxembourg, Luxembourg, 
 \textsuperscript{2}LMU Munich, Germany
\\
 \textsuperscript{3}Munich Center for Machine Learning, Germany
\\
 \textsuperscript{4}LIST, Luxembourg, 
 \textsuperscript{5}Lancaster University, UK
\\
 \small{
   \textbf{Correspondence:} \href{mailto:alistair.plum@uni.lu}{alistair.plum@uni.lu}
 }
}

\begin{document}
\maketitle
\begin{abstract}
This paper presents \textsc{ltzGLUE}, the first Natural Language Understanding (NLU) benchmark for Luxembourgish (LTZ) based on the popular GLUE benchmark for English. Although NLU tasks are available for many European languages nowadays, LTZ is one of the official national languages that is often overlooked. We construct new tasks and reuse existing ones to introduce the first official NLU benchmark and accompanying evaluation of encoder models for the language. Our tasks include common natural language processing tasks in binary and multi-class classification settings, including named entity recognition, topic classification, and intent classification. We evaluate various pre-trained language models for LTZ to present an overview of the current capabilities of these models on the LTZ language.
\vspace{-.15cm}
\end{abstract}
\section{Introduction}
\vspace{-.1cm}
\label{sec:introduction}
Language models now support Natural Language Processing (NLP) tasks in more languages than ever before \cite{10.1162/tacl_a_00365}. Advances since the introduction of the Transformer architecture \cite{vaswani2017attention, devlin-etal-2019-bert,tay2020efficient} have led to substantial performance gains, enabling Large Language Models (LLMs) to achieve state-of-the-art results across a wide range of tasks. As a result, both closed and open-weight LLMs have become the models of choice in NLP and related fields. Owing to their architecture and exposure to large-scale multilingual pre-training data, these models often demonstrate strong performance across many languages. Moreover, their ability to be fine-tuned for a wide variety of downstream tasks enhances their multilingual capabilities.

The perceived support for a wide range of languages has created an unprecedented need for language-specific evaluation of language models. As access to LLMs becomes increasingly widespread, so too does the belief that these models perform well across all languages, an assumption that does not always hold in practice \cite{zhang2023donttrustchatgptquestion}. In the interest of transparency and responsible deployment, it is therefore essential to systematically evaluate the Natural Language Understanding (NLU) capabilities of language models \cite{hettiarachchi-etal-2026-overview}.

Small and under-researched languages are particularly difficult to evaluate, as is the case with Luxembourgish (LTZ), the national language of Luxembourg, with around 400k speakers. In English, multiple benchmarks for NLU exist, including GLUE \cite{wang-etal-2018-glue} and SuperGLUE \cite{wang19superglue}, and more for other large and small languages alike \cite{park2021klue,basile-etal-2023-uinauil,hardalov-etal-2023-bgglue,shavrina-etal-2020-russiansuperglue}. However, this is not the case for LTZ, which only has a handful of NLU tasks available \cite{lothritz-etal-2022-luxembert,philippy2024,plum2026ner}. As most of these are in the news domain, and the majority of the down-stream tasks comprise less than a thousand instances, model evaluation is not always dependable. Additional factors, such as the ongoing standardisation of the language \cite{Gilles_uberdachung}, vast amounts of variation \cite{lutgen2025}, and decentralised resources, make it extremely challenging to evaluate LTZ language understanding in language models.

To address these gaps, we introduce \textsc{ltzGLUE}, a general language understanding benchmark for LTZ that includes new and existing NLU tasks. Moreover, we evaluate various pre-trained language models for LTZ to ascertain the current state of the art for the language. Our contributions are:
\begin{itemize}
    \setlength{\itemsep}{0pt}
    \setlength{\parskip}{0pt}
    \setlength{\parsep}{0pt}
    \item[(1)] \textsc{ltzGLUE}: the first unified GLUE benchmark for LTZ, with 8 tasks.\footnote{\url{https://github.com/plumaj/ltzGLUE}}
    \item[(2)] \modelname (mini/base): 2 new encoder language models for LTZ, which achieve competitive performance when fine-tuned on \textsc{ltzGLUE}.\footnote{\url{https://huggingface.co/instilux}}
    \item[(3)] A systematic evaluation of new and existing models for LTZ.
\end{itemize}

\section{Related Work}
\label{sec:rel}
Work on language understanding has progressed along two largely separate lines: large-scale benchmarking for high-resource languages and emerging efforts to build resources for smaller ones. The first has produced influential frameworks which have shaped evaluation practices for pre-trained models across domains. The second has focused on adapting NLP methods to under-researched languages, where data scarcity and linguistic variation remain major challenges. 

The General Language Understanding Evaluation (GLUE) benchmark \cite{wang-etal-2018-glue} became a cornerstone of NLU research by consolidating diverse tasks such as sentiment analysis \cite{socher2013recursive}, textual entailment \cite{williams2018broad}, and paraphrase detection \cite{dolan2005automatically} into a unified evaluation framework. It established a shared reference point for pre-trained models like \textsc{BERT} \cite{devlin-etal-2019-bert}, \textsc{RoBERTa} \cite{liu2019roberta}, and \textsc{XLNet} \cite{yang2019xlnet}, and allowed for systematic comparison across architectures and training regimes. 
\textsc{SuperGLUE} \cite{wang19superglue} addressed some shortcomings by introducing more challenging tasks such as \textsc{COPA} \cite{roemmele2011choice}, \textsc{WSC} \cite{levesque2012winograd}, and \textsc{MultiRC} \cite{khashabi2018looking}, shifting focus toward commonsense inference and multi-sentence reasoning. While both benchmarks remained English-only, the methodological influence extended widely, shaping later evaluation design in terms of robustness, transparency, and reproducibility. 

Multilingual benchmarks in the GLUE style have also been developed, including \textsc{XGLUE} \cite{liang2020xglue}, \textsc{XTREME} \cite{hu2020xtreme}, and \textsc{XTREME-R} \cite{ruder2021xtremer}, as well as language-specific adaptations such as \textsc{KLUE} \cite{park2021klue}, \textsc{RussianSuperGLUE} \cite{shavrina-etal-2020-russiansuperglue}, \textsc{bgGLUE} \cite{hardalov-etal-2023-bgglue} and \textsc{sinhalaGlue} \cite{ranasinghe-etal-2025-sinhala}. These efforts highlighted the limits of cross-lingual transfer, reinforcing the need for careful, language-specific evaluation beyond high-resource settings.

\subsection{Luxembourgish NLP}
\label{sec:rel_lux}
LTZ, the focus of this benchmark, is regarded as under-researched, and research is ongoing. \citet{joshi2020state} classify Luxembourgish as one of the ``scraping-by'' languages: although some unlabeled data exists, meaningful progress will require coordinated efforts to raise awareness and collect labeled datasets, as such resources are currently almost nonexistent. Nevertheless, the first computational tools and corpora were introduced by \citet{adda-decker-etal-2008-developments}, followed by orthographic studies such as contextual \textit{n}-deletion in transcribed speech \cite{snoeren-etal-2010-study}. \citet{lavergne-etal-2014-automatic} later provided one of the earliest annotated datasets for mixed-language processing. These efforts, though limited, established the foundations for subsequent large-scale data creation and model development.

Since then, the range of tasks has expanded considerably. Work on sentiment analysis \cite{sirajzade-etal-2020-sentiment, Gierschek2022}, orthographic correction \cite{purschke2020attitudes}, and syntactic annotation \cite{plum2024} has broadened the empirical basis for LTZ NLP. Additional datasets have targeted topic classification \cite{philippy2024}, comment moderation \cite{ranasinghe-etal-2023-publish}, and orthographic normalisation \cite{lutgen2025}, alongside the generative benchmark set \textsc{LuxGen} \cite{plum2025}. A manually annotated classification resource was introduced by \cite{lothritz-etal-2022-luxembert}, which covers a variety of classification tasks.

Model development has explored different transfer and training strategies. \textsc{LuxGPT} \cite{bernardy2022} applied cross-lingual transfer from German, \textsc{LuxemBERT} \cite{lothritz-etal-2022-luxembert} used data augmentation with generated samples, and \textsc{LuxT5} \cite{plum2025} extended multilingual pre-training with a balanced language representation. 

Yet progress remains uneven across tasks, and existing resources vary widely in size, domain, and annotation quality. No unified benchmark currently exists to evaluate LTZ language understanding consistently, a gap we aim to fill.

\section{Tasks}\label{sec:tasks}
In this section, we introduce the eight tasks for \textsc{ltzGLUE}. 
The set spans binary and multi-class sentence and token-level classification tasks. Together, these tasks cover a broad spectrum of linguistic and semantic phenomena and provide the first unified benchmark for evaluating LTZ NLP models. See Table \ref{tab:task_examples} in the Appendix for task examples.

Unless stated otherwise, the textual data used across most tasks stems from two main sources: (i) RTL\footnote{\url{https://rtl.lu}} is the main news broadcaster in Luxembourg, and the only one that is completely in LTZ. RTL provides news articles from the time span 2008 until 2024. (ii) Wikipedia has a growing set of articles in LTZ, around 66k at the time of writing.

\subsection{Headline Acceptability} 
We formulate headline acceptability (\hc) as a binary classification task where the model must decide whether a given headline matches the accompanying article body. To construct this dataset, we use RTL news articles. We keep only documents from the twenty most frequent categories. We then filter articles by body length and title length, remove exact duplicate titles, randomly shuffle the remaining instances, and retain a fixed subset of 30k examples. This subset is split equally, with one half serving as the positive class with original headlines, and the other half providing the article bodies for which we assign swapped headlines.

The swapping itself is based on a document level similarity space constructed over the full corpus. We compute TF–IDF representations of the article texts using unigrams and bigrams, an LTZ stopword list, a minimum document frequency of two, and a large feature cap to preserve topical detail. On this representation, we build a cosine nearest neighbour index that returns the 100 most similar articles for any given source. In parallel, we derive a set of content tokens for each article by extracting tokens of length 3$+$, and removing stopwords. This set is used as a proxy for the article topic. The size of the intersection between two content token sets gives a simple but effective measure of topic overlap.

For every article body in the negative half, we search its nearest neighbours to identify a donor headline, with a minimum 30-day distance so that we avoid headlines tied to the same event. We score candidates by their word overlap, which is computed as the intersection of content-word sets, and use cosine distance as a secondary tiebreaker, stopping early when the overlap reaches at least five tokens. To prevent trivial matches, we reject candidates whose headlines show high positional similarity, measured as the fraction of identical tokens in aligned positions (threshold 0.25). If no neighbour passes all criteria, we fall back to the first viable option, or ultimately to the first non-identical neighbour. We store original and swapped titles, reshuffle, and split into train (20k), development (3k), and test (6k) sets. The resulting negative examples remain topically related but are temporally and structurally mismatched, forcing models to attend to article content rather than surface cues.

\subsection{Sentiment Analysis}
We formulate the sentiment analysis (\sa) task as a classification task where the model has to predict \textit{positive}, \textit{negative}, and \textit{neutral} sentiment. We use articles from RTL, randomly selected from the \textit{commentary} and \textit{letter to the editor} sections. We chose these two specific sections since these pieces could be written by every reader of the journal or expert of a given topic and are usually comments to national or international events. Therefore, there is no required objectivity or impartiality in the writing.

In total, we extract 4,583 sentences, which are then annotated by two native speakers of LTZ. Annotators are instructed to label each sentence, and to use \textit{unsure} only when they would otherwise randomly use the other labels. We calculated Cohen's Kappa at 0.45. For the final set, the annotators agree on a label in cases of label disagreement.

\begin{table}[ht!]
\centering
\small
\begin{tabular}{lrrr}
\toprule
\textbf{Sentiment} & \textbf{Train} & \textbf{Dev} & \textbf{Test} \\
\midrule
Neutral  & 2{,}339 & 382 & 679 \\
Positive & 136     & 43  & 54  \\
Negative & 547     & 172 & 193 \\
\midrule
\textbf{Total} & 3{,}022 & 597 & 926 \\
\bottomrule
\end{tabular}
\caption{Sentiment label distribution per split.}
\label{tab:sentiment-counts}
\end{table}

\subsection{Linguistic Acceptability} 
We introduce a linguistic acceptability dataset consisting of four distinct linguistic subtypes, which can either be used as a binary (\lab) or multiclass  (\lam) classification dataset. The sentences are derived from the Luxembourgish Online Dictionary (LOD) and are manipulated using the tags available in the dataset.\footnote{\url{https://lod.lu}} 

The first class interferes with the subject-verb agreement by changing the conjugated form of the main verb or auxiliary verb. The second class similarly modifies the declined form of the adjective and therefore violates the agreement in case, number, and gender. For the third class, we manipulate the syntax by deleting 2-3 random words from the sentence, depending on the length. The last class impacts the orthography, which is achieved by using data provided by \textit{Spellchecker.lu},\footnote{\url{https://spellchecker.lu}} a semi-automatic spellchecking website frequently used in Luxembourg. We changed one random word in the sentence by using the least frequent variant in the spellchecker data. The multiclass dataset and binary dataset have a 70-10-20 split, and the distribution is shown in Table \ref{tab:ling_acc_mult}. The binary dataset distinguishes between correct (1) and incorrect (0), for which the label 0 encompasses the categories Verb, Adj, Syntax and Ortho. 

\begin{table}[ht]
\centering
\small
\begin{tabular}{lrrr}
\toprule
\textbf{Category} & \textbf{Train} & \textbf{Dev} & \textbf{Test} \\
\midrule
Verb     & 2{,}969 & 709 & 405 \\
Adj    & 2{,}388 & 357 & 673 \\
Syntax   & 2{,}327 & 333 & 664 \\
Ortho & 2{,}328 & 333 & 666 \\
Correct    & 4666 & 666 & 1333 \\
\midrule
\textbf{Total} & 14{,}678 & 2{,}094 & 4{,}045 \\
\bottomrule
\end{tabular}
\caption{Linguistic acceptability categories per split.}
\label{tab:ling_acc_mult}
\end{table}

\subsection{Named Entity Recognition} 
The \textsc{JudgeWEL} dataset \cite{plum2026ner} introduces an automatically constructed corpus for named entity recognition (\ner) in LTZ, derived from Wikipedia and Wikidata. Using Wikipedia's hyperlink structure, entities are matched to their corresponding Wikidata types and labelled in BIO format. Candidate sentences are selected to maximise diversity, and a set of quality heuristics filters incomplete or overlapping entities. The resulting sentences are then evaluated using LLMs acting as judges, with minimal human verification to calibrate quality thresholds. The final dataset contains roughly 27k sentences across five entity types (see Table \ref{tab:ner-counts}). Models trained on \textsc{JudgeWEL} achieve performance comparable to human-annotated data, demonstrating that automatically constructed resources can provide effective supervision.

The \ner dataset introduced by \citet{lothritz-etal-2022-luxembert}, by contrast, is a fully human-annotated corpus derived from RTL online news comments. It covers a wider range of text types and registers, including informal and code-mixed writing, and focuses on four primary entity categories (PER, ORG, LOC, GPE). Annotation was conducted manually, yielding a smaller but high-precision dataset. 

The two datasets are merged to increase both coverage and domain balance. To ensure compatibility, the tag set is harmonised by merging the GPE and LOC categories into a single location label, while retaining PER, ORG, and MISC unchanged. This unified resource thus aligns the structured reliability of \textsc{JudgeWEL} with the domain and stylistic breadth of the \ner set by \cite{lothritz-etal-2022-luxembert}, providing a large-scale, multi-domain \ner dataset for LTZ. See entity type counts in Table \ref{tab:ner-counts}. 

\begin{table}[ht!]
\centering
\small
\begin{tabular}{lrrr}
\toprule
\textbf{Entity Type} & \textbf{Train} & \textbf{Dev} & \textbf{Test} \\
\midrule
PER  & 11{,}961 & 1{,}587 & 1{,}449 \\
ORG  & 3{,}323  & 423     & 385     \\
LOC  & 11{,}701 & 1{,}503 & 1{,}425 \\
DATE & 11{,}355 & 1{,}414 & 1{,}523 \\
MISC & 511      & 116     & 40      \\
\midrule
\textbf{Total} & 38{,}851 & 5{,}043  & 4{,}822 \\
\bottomrule
\end{tabular}
\caption{Entity type counts per split.}
\label{tab:ner-counts}
\end{table}

\subsection{Topic Classification} 
To construct the news topic classification (\tc) dataset, we collected news articles from RTL, which provides content pre-assigned to editorial categories. We applied a series of preprocessing steps to ensure data quality. Specifically, we removed articles identified as non-Luxembourgish by OpenLID \cite{burchell-etal-2023-open}, as well as those containing fewer than 40 words or more than 400 words. From the available categories, we focused on five principal domains: \textsc{sports}, \textsc{culture}, \textsc{technology}, \textsc{business}, and \textsc{animals}. Given the substantial over-representation of the \textsc{sports} category, we performed downsampling to mitigate class imbalance. The resulting dataset was split into training, development, and test sets (category distribution is summarized in Table~\ref{tab:dataset-splits}).

\begin{table}[ht]
\centering
\small
\begin{tabular}{lrrr}
\toprule
\textbf{Category} & \textbf{Train} & \textbf{Dev} & \textbf{Test} \\
\midrule
Sports     & 4{,}000 & 500 & 500 \\
Culture    & 2{,}984 & 373 & 374 \\
Business   & 1{,}111 & 138 & 140 \\
Technology & 1{,}027 & 128 & 129 \\
Animals    & 810 & 101 & 102 \\
\midrule
\textbf{Total} & 9{,}932 & 1{,}240 & 1{,}245 \\
\bottomrule
\end{tabular}
\caption{News topics per split.}
\label{tab:dataset-splits}
\end{table}

\subsection{Intent Detection} 
We constructed a new LTZ dataset for intent detection (\sid) by translating the English xSID test and validation datasets \cite{van-der-goot-etal-2021-masked}. The translations were performed by an LTZ native speaker. In cases of uncertainty, additional native LTZ speakers were consulted. Since LTZ is linguistically closely related to German, the German dataset \cite{van-der-goot-etal-2021-masked} occasionally served as a reference point. Since this task is originally intended to be crosslingual, we use the machine translated German training set \cite{van-der-goot-etal-2021-masked}.

The main challenge in translating the English dataset stems from its register. The source segments consist of user commands for a voice-controlled AI assistant, representing a specialised spoken register for which there is no equivalent reference corpus in LTZ. This register is marked by domain-specific terminology and collocations (e.g., \textit{set an alarm}, \textit{set a reminder}, \textit{add to playlist}), as well as non-standard spelling (e.g., all lower-case, missing punctuation). Due to the lack of LTZ references in this register, it was not possible to systematically verify the translated terminology.


After translating the dataset, we transferred the BIO tags by first using token-level fuzzy matching between the LTZ and the German dataset, followed by manual verification. Table \ref{tab:intent-splits} shows the label distribution and size of each data split.

\begin{table}[ht]
\centering
\small
\begin{tabular}{lrrr}
\toprule
\textbf{Category} & \textbf{Train} & \textbf{Dev} & \textbf{Test} \\
\midrule
AddToPlaylist            & 1{,}842 & 1 & 2 \\
BookRestaurant           & 1{,}873 & 8 & 11 \\
PlayMusic                & 1{,}900 & 3 & 5 \\
RateBook                 & 1{,}856 & 4 & 3 \\
SearchCreativeWork       & 1{,}854 & 0 & 9 \\
SearchScreeningEvent     & 1{,}859 & 6 & 4 \\
alarm/cancel\_alarm      & 2{,}069 & 0 & 1 \\
alarm/modify\_alarm      & 439     & 0 & 0 \\
alarm/set\_alarm         & 4{,}816 & 4 & 4 \\
alarm/show\_alarms       & 1{,}142 & 1 & 0 \\
alarm/snooze\_alarm      & 432     & 0 & 0 \\
alarm/time\_left\_on\_alarm & 384  & 0 & 0 \\
reminder/cancel\_reminder & 1{,}151 & 0 & 0 \\
reminder/set\_reminder   & 4{,}743 & 1 & 3 \\
reminder/show\_reminders & 1{,}006 & 0 & 0 \\
weather/checkSunrise     & 124     & 0 & 0 \\
weather/checkSunset      & 168     & 0 & 0 \\
weather/find             & 15{,}947 & 25 & 24 \\
\midrule
\textbf{Total}           & 36{,}605 & 53 & 66 \\
\bottomrule
\end{tabular}
\caption{Intent distribution per data split.}
\label{tab:intent-splits}
\end{table}

\subsection{Recognizing Textual Entailment} 
Recognizing Textual Entailment (\rte)~\cite{haim2006second} is a classic NLU task featured in the original GLUE benchmark. Given a pair of texts A and B, the task consists of determining whether A is a logical premise of B. \citet{lothritz2023comparing} released a machine-translated Luxembourgish version of the dataset using Google Translate. However, due to numerous grammar and vocabulary related mistakes introduced in this process, we set out to improve the quality of the dataset.

Specifically, we first prompted \textsc{ChatGPT-5.1} to assess and improve the translated sentence pairs unless they were already of very high quality, while explicitly keeping the original meaning to avoid label conflicts (see Appendix~\ref{app:improve_rte}). In addition, we perform two verification steps to make sure that (a) the quality of the improved texts is high enough and (b) that the labels are correct. 

To achieve (a), we prompted \textsc{ChatGPT-5-mini} to judge the texts in the improved data and label their quality as either \textit{low}, \textit{medium}, or \textit{high}, keeping only data rated at least \textit{medium}, removing nearly 25\% of the entire dataset (see Appendix~\ref{app:judge_rte}).

For (b), we prompted \textsc{ChatGPT-5-mini} to verify whether the dataset labels remained correct after the first translation and improvement, outputting \textit{true} or \textit{false} for each sentence pair (see Appendix~\ref{app:verify_rte_labels}). 
Nearly 10\% of the labels were \textit{false}. We found that the quality improvement step often corrected intentional logical contradictions or factual inaccuracies rather than keeping the original semantics. We therefore adjusted the sentences manually such that they corresponded to the ground truth again, while keeping false positives intact.


The filtering reduced between 22 and 28\% of instances in the data, resulting in a final dataset of 1,876, 197, and 626 sentence pairs for the training, development, and test set, respectively.

\subsection{Summary}
Together, the eight tasks in \textsc{ltzGLUE} form a broad and balanced evaluation suite, covering four binary and four multi-class settings, sentence- and document-level inputs, as well as a token-level sequence-labelling task. Despite the low-research status of LTZ, this places \textsc{ltzGLUE} in the same general range as the original English GLUE benchmark, which comprises nine diverse NLU tasks \citep{wang-etal-2018-glue}. In addition, a substantial proportion of the \textsc{ltzGLUE} tasks are newly created for LTZ rather than direct translations or simple repackaging, allowing the benchmark to reflect phenomena and usage patterns specific to the language.

Compared to recent GLUE-style benchmarks for other non-English languages, \textsc{ltzGLUE} also offers competitive, and in some respects stronger, task coverage. \textsc{Sinhala-GLUE}, introduced as part of the Sinhala encoder-only language models and evaluation suite \citep{ranasinghe-etal-2025-sinhala}, bundles six datasets into a single NLU benchmark, while \textsc{UINAUIL} provides six harmonised Italian NLU tasks drawn from existing shared-task resources \citep{basile-etal-2023-uinauil}. For Bulgarian, \textsc{bgGLUE} defines nine NLU tasks, combining sequence labelling, document-level classification, and regression over established datasets \citep{hardalov-etal-2023-bgglue}. In this landscape, supporting eight tasks for LTZ, including token-level \ner and several newly constructed text-level tasks, is a strong indicator of the maturity and breadth of the emerging LTZ NLP ecosystem.


\section{Models}\label{sec:models}
This section presents the models we trained and evaluated with \textsc{ltzGLUE}. We cover both supervised encoder-based architectures fine-tuned on the benchmark tasks and prompt-based large language models. This design allows us to assess current LTZ NLU performance across fundamentally different modelling paradigms, while maintaining a clear separation between task-specific supervision and general-purpose language understanding.

\subsection{\modelname}\label{sec:model}
We train two encoder language models for LTZ: \modelname-mini with 68M and \modelname-base with 110M non-embedding parameters. We closely follow the Ettin recipe \cite{weller2025seqvsseqopen}, which is based on \textsc{ModernBERT} \cite{warner2024smarterbetterfasterlonger} (see Appendix \ref{app:arch} for detailed settings). 

The pre-training set is compiled from a variety of sources of LTZ. A large portion of the data stems from RTL (see Section \ref{sec:tasks}), including news articles (News), transcribed radio interviews (Radio), and user comments (Comments). We also include transcribed podcasts (Podcasts) and transcribed political speeches and debates from the Chambre des Députés (Chamber). In addition, we use 1M sentences from the web crawl of the Leipzig Collection (Web, this excludes RTL), text crawled from LTZ chat rooms (Webchat), a Wikipedia crawl from October 2023 (Wikipedia), and finally, example sentences from the LOD retrieved in March 2024. We filter out sentences containing fewer than three words (as tokenized by whitespace), totalling 11.7M sentences, which corresponds to roughly 233M tokens using our tokenizer. Token counts per source can be found in \cref{tab:pre-training-counts} in the Appendix.

\subsection{Supervised}
We evaluate a set of supervised encoder-based models that explicitly support LTZ, either through direct pre-training or multilingual coverage. As a representative baseline, we include multilingual BERT (\mbert) \cite{devlin-etal-2019-bert}, which still remains widely used for multilingual transfer and low-resource evaluation. We additionally evaluate a more recent multilingual BERT (\mmbert) variant with updated pre-training data and tokenisation. 

To complement these general-purpose multilingual models, we include \luxembert, a language-specific model trained on LTZ data \cite{lothritz-etal-2022-luxembert}, which provides a stronger inductive bias for the language’s lexical and orthographic properties. Finally, we evaluate XLM-RoBERTa (\roberta) \cite{conneau2019}, a large-scale multilingual model trained on substantially more data and languages than \mbert, and commonly used as a strong reference point for multilingual NLU. 

\subsection{Unsupervised}
In addition to supervised encoder-based models, we evaluate a set of LLMs in a prompt-based zero-shot setting. This group includes \textsc{Qwen3-235B}, \textsc{LLaMA-3.3}, \textsc{Gemma-3-27B}, and \textsc{GPT-5-mini}, which represent a range of model sizes, training regimes, and degrees of multilingual coverage. None of these models are fine-tuned on \textsc{ltzGLUE}, although some of the text data (RTL, Wikipedia) is very likely to have been processed during training. The models are evaluated using prompts that describe each task, allowing us to assess their ability to generalise to LTZ without task-specific supervision (see Appendix \ref{app:zeroshot_llms} and \ref{app:zeroshot_llms_tasks} for further details). We did not use a Multiple Choice Question Answering (MCQA)-setup, but provided the labels that should be used as output. 

This evaluation setting reflects the growing use of LLMs as general-purpose language understanding systems, particularly in scenarios where annotated data is scarce or unavailable. However, prompt-based evaluation introduces additional sources of variability, including prompt sensitivity and differences in instruction-following behaviour across models. As a result, performance should be interpreted as indicative rather than directly comparable to supervised results. Nevertheless, including these models provides a complementary perspective on the current capabilities of large-scale multilingual and instruction-tuned systems for LTZ NLU.

\section{Evaluation}\label{sec:evaluation}
We evaluate the models described in Section \ref{sec:models} across all tasks in the benchmark. For encoder-based models, results are reported as averages over multiple runs (see Appendix \ref{app:arch} for more details). Prompted LLMs do not always produce well-formed outputs and may return an incorrect number of predictions for a given task; such outputs are discarded prior to evaluation. All reported scores are computed on the remaining valid predictions per model. For the supervised models, since the linguistic acceptability and sentiment analysis datasets are highly imbalanced, when fine-tuning on these tasks we use class-balanced loss based on effective size \cite{cui2019classbalancedlossbasedeffective} with a beta of 0.99. Table \ref{tab:all_res} shows \Fone scores for all models across all tasks (see Appendix \ref{app:full} for full results).

Overall, our evaluation reveals consistent trends across tasks. Encoder-based models perform strongly across most settings, particularly on structurally complex and label-sensitive tasks, confirming findings from prior work on multilingual and low-resource NLU \citep{wu-dredze-2019-beto,conneau2019}. 
Prompted large language models, by contrast, show more variable behaviour and perform competitively only on a set of semantically coarse-grained tasks, consistent with recent observations that prompting alone is often insufficient for strong performance on structured NLU tasks \citep{wei-etal-2022-chain,liu-etal-2023-pretrain}. 

\newcommand{\foneoneshot}[1]{#1}

\begin{table*}[t]
\centering
\small
\begin{tabular}{lcccccccc}
\toprule
\textbf{Model} 
& \textbf{\hc} 
& \textbf{\sa} 
& \textbf{\labs} 
& \textbf{\lams} 
& \textbf{\ner} 
& \textbf{\tc} 
& \textbf{\sid} 
& \textbf{\rte} \\
\midrule

\luxemberts
& \cellcolor[HTML]{CBEADB}\fonesd{66.37}{0.00}
& \cellcolor[HTML]{57BB8A}\fonesd{\textbf{58.66}}{0.73}
& \cellcolor[HTML]{6EC59A}\fonesd{89.17}{0.18}
& \cellcolor[HTML]{6EC59A}\fonesd{87.96}{0.71}
& \cellcolor[HTML]{BEE5D3}\fonesd{87.43}{1.22}
& \cellcolor[HTML]{72C69D}\fonesd{98.68}{0.17}
& \cellcolor[HTML]{57BB8A}\fonesd{\textbf{91.71}}{0.11}
& \cellcolor[HTML]{DCF1E7}\fonesd{46.51}{5.16} \\

\mberts
& \cellcolor[HTML]{72C69D}\fonesd{77.91}{10.26}
& \cellcolor[HTML]{DBF1E6}\fonesd{41.25}{4.87}
& \cellcolor[HTML]{A0D9BD}\fonesd{81.26}{0.56}
& \cellcolor[HTML]{A0D9BD}\fonesd{81.20}{1.02}
& \cellcolor[HTML]{BEE5D3}\fonesd{83.06}{2.06}
& \cellcolor[HTML]{BEE5D3}\fonesd{97.80}{0.67}
& \cellcolor[HTML]{DCF1E7}\fonesd{60.65}{4.71}
& \cellcolor[HTML]{E5F5ED}\fonesd{42.57}{6.13} \\

\modelnames
& \cellcolor[HTML]{DCF1E7}\fonesd{62.81}{4.98}
& \cellcolor[HTML]{B3E0CA}\fonesd{46.59}{5.88}
& \cellcolor[HTML]{A0D9BD}\fonesd{83.17}{8.38}
& \cellcolor[HTML]{BEE5D3}\fonesd{78.63}{11.55}
& \cellcolor[HTML]{BEE5D3}\fonesd{88.01}{1.07}
& \cellcolor[HTML]{57BB8A}\fonesd{\textbf{98.95}}{0.36}
& \cellcolor[HTML]{A0D9BD}\fonesd{73.32}{11.66}
& \cellcolor[HTML]{DCF1E7}\fonesd{39.38}{6.04} \\

\modelnamesm
& \cellcolor[HTML]{BEE5D3}\fonesd{72.69}{1.33}
& \cellcolor[HTML]{BCE4D0}\fonesd{45.39}{6.79}
& \cellcolor[HTML]{6EC59A}\fonesd{89.31}{2.89}
& \cellcolor[HTML]{6EC59A}\fonesd{86.62}{4.64}
& \cellcolor[HTML]{BEE5D3}\fonesd{88.95}{0.61}
& \cellcolor[HTML]{72C69D}\fonesd{98.50}{0.23}
& \cellcolor[HTML]{96D5B6}\fonesd{80.13}{3.00}
& \cellcolor[HTML]{E5F5ED}\fonesd{45.35}{5.71} \\

\mmberts
& \cellcolor[HTML]{57BB8A}\fonesd{\textbf{85.59}}{1.61}
& \cellcolor[HTML]{7FCCA6}\fonesd{53.37}{4.47}
& \cellcolor[HTML]{57BB8A}\fonesd{\textbf{89.97}}{0.09}
& \cellcolor[HTML]{57BB8A}\fonesd{\textbf{88.83}}{1.23}
& \cellcolor[HTML]{57BB8A}\fonesd{\textbf{90.41}}{0.55}
& \cellcolor[HTML]{72C69D}\fonesd{98.92}{0.28}
& \cellcolor[HTML]{A0D9BD}\fonesd{78.26}{7.22}
& \cellcolor[HTML]{72C69D}\fonesd{\textbf{52.81}}{3.01} \\

\robertas
& \cellcolor[HTML]{BEE5D3}\fonesd{72.09}{9.90}
& \cellcolor[HTML]{FFFFFF}\fonesd{36.40}{0.57}
& \cellcolor[HTML]{DCF1E7}\fonesd{70.25}{4.55}
& \cellcolor[HTML]{DCF1E7}\fonesd{73.40}{5.61}
& \cellcolor[HTML]{BEE5D3}\fonesd{79.58}{2.77}
& \cellcolor[HTML]{BEE5D3}\fonesd{97.24}{0.50}
& \cellcolor[HTML]{DCF1E7}\fonesd{62.75}{8.76}
& \cellcolor[HTML]{DCF1E7}\fonesd{35.24}{7.73} \\

\midrule

\textsc{GPT}
& \cellcolor[HTML]{57BB8A}\foneoneshot{\textbf{88.88}}
& \cellcolor[HTML]{A0D9BD}\foneoneshot{\textbf{56.44}}
& \cellcolor[HTML]{72C69D}\foneoneshot{\textbf{75.24}}
& \cellcolor[HTML]{BEE5D3}\foneoneshot{\textbf{51.45}}
& \cellcolor[HTML]{A0D9BD}\foneoneshot{67.15}
& \cellcolor[HTML]{72C69D}\foneoneshot{\textbf{89.27}}
& \cellcolor[HTML]{DCF1E7}\foneoneshot{\textbf{38.26}}
& \cellcolor[HTML]{57BB8A}\foneoneshot{\textbf{88.51}} \\

\textsc{Qwen}
& \cellcolor[HTML]{72C69D}\foneoneshot{86.08}
& \cellcolor[HTML]{DCF1E7}\foneoneshot{51.63}
& \cellcolor[HTML]{BEE5D3}\foneoneshot{67.77}
& \cellcolor[HTML]{DCF1E7}\foneoneshot{39.69}
& \cellcolor[HTML]{72C69D}\foneoneshot{\textbf{70.73}}
& \cellcolor[HTML]{72C69D}\foneoneshot{88.37}
& \cellcolor[HTML]{DCF1E7}\foneoneshot{34.90}
& \cellcolor[HTML]{BEE5D3}\foneoneshot{84.17} \\

\textsc{Gemma}
& \cellcolor[HTML]{BEE5D3}\foneoneshot{80.53}
& \cellcolor[HTML]{DCF1E7}\foneoneshot{55.60}
& \cellcolor[HTML]{DCF1E7}\foneoneshot{58.68}
& \cellcolor[HTML]{DCF1E7}\foneoneshot{43.28}
& \cellcolor[HTML]{DCF1E7}\foneoneshot{48.44}
& \cellcolor[HTML]{DCF1E7}\foneoneshot{48.67}
& \cellcolor[HTML]{FFFFFF}\foneoneshot{7.25}
& \cellcolor[HTML]{BEE5D3}\foneoneshot{73.12} \\

\textsc{Llama}
& \cellcolor[HTML]{BEE5D3}\foneoneshot{77.66}
& \cellcolor[HTML]{DCF1E7}\foneoneshot{45.84}
& \cellcolor[HTML]{DCF1E7}\foneoneshot{41.44}
& \cellcolor[HTML]{FFFFFF}\foneoneshot{11.17}
& \cellcolor[HTML]{DCF1E7}\foneoneshot{50.50}
& \cellcolor[HTML]{FFFFFF}\foneoneshot{12.47}
& \cellcolor[HTML]{DCF1E7}\foneoneshot{35.75}
& \cellcolor[HTML]{BEE5D3}\foneoneshot{72.01} \\

\bottomrule
\end{tabular}
\caption{\textbf{Test \Fone scores across all ltzGLUE tasks.}
Encoder results are averaged over three runs with standard deviations as subscripts. Prompted LLMs were evaluated once; we report macro-\Fone only.
}
\label{tab:all_res}
\end{table*}

\paragraph{\hc}
Results on the headline acceptability task show substantial variation across encoder-based models, both in absolute performance and in stability. \mmbert achieves the highest mean \Fone score with comparatively low variance, indicating robust performance across runs. In contrast, \mbert reaches competitive average performance but exhibits very high standard deviation, suggesting sensitivity to initialisation and training dynamics. The LTZ-specific encoders, \luxembert and \modelname-mini, perform moderately well but remain clearly below \mmbert, while \modelname-base and \roberta lag behind in both performance and consistency. Among the prompted LLMs, \textsc{GPT} achieves the strongest single-run result, approaching the performance of \mmbert, followed by \textsc{Qwen}. \textsc{Gemma} and \textsc{Llama} perform noticeably worse. However, these results are based on a single evaluation and therefore do not allow conclusions about stability.

\paragraph{\sa}
On the sentiment analysis task, differences between encoder models are smaller than for \hc, though consistent trends remain. \luxembert achieves the highest mean \Fone score with low variance, followed closely by \mmbert, although with considerable variance across runs.  \modelname-base, \modelname-mini, and \mbert perform worse and exhibit increased variability, while \roberta performs weakest among the encoders. Prompted LLMs perform roughly equal to the fine-tuned encoders in this setting. \textsc{GPT} achieves the strongest single-run \Fone score among the LLMs, marginally outperforming \textsc{Gemma}.

\paragraph{\lab}
For the binary linguistic acceptability task, all encoder models achieve relatively high \Fone scores, with \mmbert and \modelname-mini performing best and showing limited variance across runs. \luxembert also performs competitively, suggesting that coarse-grained acceptability judgments are well captured by language-specific representations. In contrast, \modelname-base exhibits notably higher variance despite a reasonable mean score, complicating direct comparison. \roberta performs substantially worse than the other encoders. Prompted LLMs trail the encoder models by a clear margin: \textsc{GPT} achieves the highest single-run performance, followed by \textsc{Qwen}, while \textsc{Gemma} and \textsc{Llama} perform considerably worse. These results indicate that even binary acceptability judgments benefit from task-specific supervision.

\paragraph{\lam}
The multi-class linguistic acceptability task proves considerably more challenging and reveals larger performance differences. Among the encoders, \mmbert again leads, combining strong performance with moderate variance. \luxembert and \modelname-mini follow closely but show increased instability across runs, while \modelname-base exhibits particularly high standard deviation, suggesting difficulty in consistently modeling fine-grained acceptability distinctions. \mbert performs slightly worse than the LTZ-specific encoders, and \roberta remains the weakest. Prompted LLM performance drops sharply in this setting: although \textsc{GPT} achieves the highest single-run score, all LLMs perform well below the encoders, with \textsc{Llama} approaching chance-level behaviour. This highlights the difficulty of multi-class linguistic judgments without supervised adaptation.

\paragraph{\ner}
Results on the named entity recognition task show strong performance across all encoder-based models, with comparatively small differences in mean \Fone scores. \mmbert achieves the highest score with very low variance, indicating both high accuracy and stability. \modelname-mini and \modelname-base perform similarly well, while \luxembert remains competitive but slightly behind. \mbert and \roberta trail the other encoders. In contrast, prompted LLMs perform substantially worse than all fine-tuned encoders. \textsc{Qwen} achieves the strongest LLM performance, followed by \textsc{GPT}, but both remain far below the encoder models, underscoring the importance of token-level supervision for this task.

\paragraph{\tc}
The topic classification task emerges as the easiest overall. All encoder models achieve very high \Fone scores with extremely low variance, indicating a stable and largely language-agnostic task. Differences between encoders are minimal, with \modelname-base and \mmbert marginally outperforming the others. Prompted LLMs perform competitively in this setting: \textsc{GPT} and \textsc{Qwen} approach encoder-level performance in a single run. However, \textsc{Gemma} and especially \textsc{Llama} perform poorly, suggesting that strong topic classification performance is not guaranteed without either fine-tuning or robust multilingual pre-training.

\paragraph{\sid}
Results on the intent detection task reveal a clear separation between models. Among the encoders, \luxembert achieves the strongest performance with very low variance, highlighting the benefit of language-specific pre-training. \modelname-mini and \mmbert perform well but exhibit higher variability, while \modelname-base shows both lower mean performance and substantial deviation across runs. \mbert and \roberta perform considerably worse. Prompted LLMs struggle substantially with this task: all LLMs achieve low \Fone scores, with \textsc{Gemma} performing particularly poorly. This suggests that intent classification in LTZ relies on supervised task-specific training.

\paragraph{\rte}
The recognising textual entailment task is the most challenging overall, with low \Fone scores and high variance across encoder models. \mmbert clearly outperforms the other encoders, achieving the highest mean performance with relatively controlled variance. \luxembert and \modelname-mini follow but show notable instability, while \modelname-base and \roberta perform poorly, making reliable inference difficult. Prompted LLMs perform relatively well in comparison to most encoders: \textsc{GPT} and \textsc{Qwen} achieve strong single-run \Fone scores, exceeding all encoder models except \mmbert. This suggests that entailment reasoning may benefit from broader semantic representations encoded in large generative models, although the lack of variance estimates warrants caution.

\paragraph{Overall}
Taken together, the results reveal three overall patterns. First, \mmbert consistently achieves the strongest or near-strongest performance across almost all tasks, combining high mean \Fone scores with comparatively low variance, suggesting that broad multilingual pre-training with sufficient LTZ exposure yields stable and transferable representations. Second, LTZ-specific encoders such as \luxembert and \modelname-mini are particularly competitive on lexically grounded or task-specific settings (e.g., intent detection and acceptability), but exhibit greater instability on structurally complex inference tasks such as multi-class acceptability and textual entailment. Third, prompted LLMs display substantially more task-dependent behaviour and generally underperform fine-tuned encoders, except on semantically coarse-grained tasks such as topic classification. Overall, tasks requiring structured prediction or fine-grained linguistic discrimination benefit strongly from supervised fine-tuning, underscoring the importance of task-specific adaptation in LTZ NLU.

\section{Conclusion}\label{sec:conclusion}
This paper makes two central contributions to LTZ NLU. First, we introduce a new benchmark that provides the first comprehensive GLUE-style evaluation suite for LTZ. Second, we present a systematic evaluation of encoder-based models and prompted large language models across all tasks, offering concrete guidance on model choice in such a low-resource setting.

The construction of the dataset required a deliberately resource-conscious approach. In the absence of large, task-diverse annotated resources, we combine the reuse of existing datasets with the targeted annotation of new data, carefully aligning annotation schemes across tasks, and using large language models as auxiliary tools. This strategy enables the creation of a benchmark without relying on large-scale annotation efforts. Moreover, our evaluation reveals a clear and consistent pattern: fine-tuned encoder-based models outperform prompted large language models on structurally complex tasks. Prompted large language models perform competitively only on a limited subset of semantically coarse-grained tasks, most notably topic classification and textual entailment. However, prompt-based approaches show limited consistency, as outputs can vary substantially across runs and prompt formulations, making prompting alone an unreliable substitute for fine-tuned models in low-resource NLU settings.

Overall, our findings indicate that, despite rapid progress in generative modelling, encoder-based approaches remain the recommended solution for most LTZ NLU tasks. Nonetheless, LLMs play an important complementary role, both as practical tools during dataset construction and as competitive baselines for selected tasks. By releasing both the dataset and the accompanying evaluation, we aim to support future research on LTZ and to encourage similarly resource-conscious benchmarking efforts for other low-resource languages.

\section*{Acknowledgments}
We would like to thank the student assistants for their annotation work.

This work is supported by the LLMs4EU project, funded by the European Union through the Digital Europe Programme (DIGITAL) under the grant agreement 10119847.
FK and BP are supported by the ERC Consolidator Grant DIALECT 101043235.

\section*{Limitations}
While \textsc{ltzGLUE} provides the first systematic benchmark for LTZ NLU, the dataset remains constrained by the availability and scope of existing resources. Several tasks rely on relatively small or domain-specific corpora, which limits the ecological validity of the results and restricts the range of linguistic phenomena covered. We therefore view this release as a foundation rather than a comprehensive evaluation suite. In addition, some of the data sources used in this benchmark may already be included, in whole or in part, in the pre-training corpora of the large language models evaluated in this work. While the exact composition of proprietary pre-training datasets is typically not fully disclosed, this potential overlap cannot be entirely ruled out and may inflate performance estimates. We explicitly acknowledge this possibility in the interest of transparency and encourage future evaluations on carefully controlled or newly collected data where feasible.

Coverage across domains, registers, and demographic varieties may also be limited. LTZ displays substantial orthographic and sociolinguistic variation, yet most data sources reflect formal writing or institutional usage and therefore do not fully represent informal and multilingual contexts. Models evaluated on \textsc{ltzGLUE} may therefore overestimate their robustness in real-world applications.


Although we draw on established GLUE-style tasks, some annotation decisions and class distributions are necessarily influenced by resource constraints. Certain tasks exhibit label imbalance or rely on automatic preprocessing, which may introduce biases that we cannot fully quantify. These constraints reflect the current state of LTZ NLP and point to the need for continued data creation and evaluation work.

\section*{Ethical Considerations}
The datasets included in this work are derived from publicly accessible sources that permit research use, and all preprocessing avoids the inclusion of directly identifying personal information. The data is licensed under the Creative Commons Attribution (CC BY) licence.

However, some tasks draw on data originally produced in institutional or media contexts, which may reflect societal biases in representation. These patterns can influence model behaviour and should be considered when deploying systems trained on \textsc{ltzGLUE}.

LTZ is a small language community, and linguistic data often originate from a limited set of public domains. As a result, models may reproduce dominant norms while under-representing regional, sociolectal, or multilingual practices. We therefore caution against using benchmark performance as evidence of cultural or demographic coverage.

Finally, although no sensitive content is intentionally included, automated filtering and preprocessing cannot guarantee the complete removal of harmful or offensive material. Researchers using \textsc{ltzGLUE} are encouraged to inspect task-specific subsets and consider downstream implications, especially in public-facing settings.

\bibliography{custom}

\begin{thebibliography}{55}
\providecommand{\natexlab}[1]{#1}

\bibitem[{Adda-Decker et~al.(2008)Adda-Decker, Pellegrini, Bilinski, and Adda}]{adda-decker-etal-2008-developments}
Martine Adda-Decker, Thomas Pellegrini, Eric Bilinski, and Gilles Adda. 2008.
\newblock {Developments of ``L{\"e}tzebuergesch{''} Resources for Automatic Speech Processing and Linguistic Studies}.
\newblock In \emph{{Proceedings of LREC}}.

\bibitem[{Basile et~al.(2023)Basile, Bioglio, Bosca, Bosco, and Patti}]{basile-etal-2023-uinauil}
Valerio Basile, Livio Bioglio, Alessio Bosca, Cristina Bosco, and Viviana Patti. 2023.
\newblock {{UINAUIL}: A Unified Benchmark for {I}talian Natural Language Understanding}.
\newblock In \emph{Proceedings of ACL}.

\bibitem[{Bernardy(2022)}]{bernardy2022}
Laura Bernardy. 2022.
\newblock {A Luxembourgish GPT-2 Approach Based on Transfer Learning}.
\newblock Master's thesis, University of Trier.

\bibitem[{Black et~al.(2022)Black, Biderman, Hallahan, Anthony, Gao, Golding, He, Leahy, McDonell, Phang, Pieler, Prashanth, Purohit, Reynolds, Tow, Wang, and Weinbach}]{black-etal-2022-gpt}
Sidney Black, Stella Biderman, Eric Hallahan, Quentin Anthony, Leo Gao, Laurence Golding, Horace He, Connor Leahy, Kyle McDonell, Jason Phang, Michael Pieler, Usvsn~Sai Prashanth, Shivanshu Purohit, Laria Reynolds, Jonathan Tow, Ben Wang, and Samuel Weinbach. 2022.
\newblock {{GPT}-{N}eo{X}-20{B}: An Open-Source Autoregressive Language Model}.
\newblock In \emph{Proceedings of BigScience}.

\bibitem[{Burchell et~al.(2023)Burchell, Birch, Bogoychev, and Heafield}]{burchell-etal-2023-open}
Laurie Burchell, Alexandra Birch, Nikolay Bogoychev, and Kenneth Heafield. 2023.
\newblock {An Open Dataset and Model for Language Identification}.
\newblock In \emph{Proceedings of ACL}.

\bibitem[{Conneau et~al.(2020)Conneau, Khandelwal, Goyal, Chaudhary, Wenzek, Guzm{\'a}n, Grave, Ott, Zettlemoyer, and Stoyanov}]{conneau2019}
Alexis Conneau, Kartikay Khandelwal, Naman Goyal, Vishrav Chaudhary, Guillaume Wenzek, Francisco Guzm{\'a}n, Edouard Grave, Myle Ott, Luke Zettlemoyer, and Veselin Stoyanov. 2020.
\newblock Unsupervised cross-lingual representation learning at scale.
\newblock In \emph{Proceedings of ACL}.

\bibitem[{Cui et~al.(2019)Cui, Jia, Lin, Song, and Belongie}]{cui2019classbalancedlossbasedeffective}
Yin Cui, Menglin Jia, Tsung-Yi Lin, Yang Song, and Serge Belongie. 2019.
\newblock Class-balanced loss based on effective number of samples.
\newblock In \emph{Proceedings of CVPR}.

\bibitem[{Devlin et~al.(2019)Devlin, Chang, Lee, and Toutanova}]{devlin-etal-2019-bert}
Jacob Devlin, Ming-Wei Chang, Kenton Lee, and Kristina Toutanova. 2019.
\newblock {BERT: Pre-training of Deep Bidirectional Transformers for Language Understanding}.
\newblock In \emph{Proceedings of NAACL-HLT}.

\bibitem[{Dolan and Brockett(2005)}]{dolan2005automatically}
William~B. Dolan and Chris Brockett. 2005.
\newblock {Automatically constructing a corpus of sentential paraphrases}.
\newblock In \emph{Proceedings of IWP}.

\bibitem[{Gierschek(2022)}]{Gierschek2022}
Daniela Gierschek. 2022.
\newblock \emph{{Detection of Sentiment in Luxembourgish User Comments}}.
\newblock Ph.D. thesis, University of Luxembourg.

\bibitem[{Gilles(2019)}]{Gilles_uberdachung}
Peter Gilles. 2019.
\newblock \emph{{39. Komplexe \"{U}berdachung II: Luxemburg. Die Genese Einer Neuen Nationalsprache}}, pages 1039--1060.
\newblock De Gruyter Mouton.

\bibitem[{Haim et~al.(2006)Haim, Dagan, Dolan, Ferro, Giampiccolo, Magnini, and Szpektor}]{haim2006second}
R~Bar Haim, Ido Dagan, Bill Dolan, Lisa Ferro, Danilo Giampiccolo, Bernardo Magnini, and Idan Szpektor. 2006.
\newblock {The second pascal recognising textual entailment challenge}.
\newblock In \emph{Proceedings of the Second PASCAL Challenges Workshop on Recognising Textual Entailment}.

\bibitem[{Hardalov et~al.(2023)Hardalov, Mihaylov, Simov, and Nakov}]{hardalov-etal-2023-bgglue}
Momchil Hardalov, Todor Mihaylov, Kiril Simov, and Preslav Nakov. 2023.
\newblock {BgGLUE: A Bulgarian General Language Understanding Evaluation Benchmark}.
\newblock In \emph{Proceedings of RANLP}.

\bibitem[{Hettiarachchi et~al.(2026)Hettiarachchi, Ranasinghe, Plum, Rayson, Mitkov, Gaber, Premasiri, Tan, and Uyangodage}]{hettiarachchi-etal-2026-overview}
Hansi Hettiarachchi, Tharindu Ranasinghe, Alistair Plum, Paul Rayson, Ruslan Mitkov, Mohamed~Medhat Gaber, Damith Premasiri, Fiona~Anting Tan, and Lasitha Uyangodage. 2026.
\newblock Overview of the second workshop on language models for low-resource languages ({L}o{R}es{LM} 2026).
\newblock In \emph{Proceedings of the {L}o{R}es{LM}}.

\bibitem[{Hu et~al.(2020)Hu, Ruder, Siddhant, Neubig, Firat, Johnson et~al.}]{hu2020xtreme}
J.~Edward Hu, Sebastian Ruder, Aditya Siddhant, Graham Neubig, Orhan Firat, Melvin Johnson, et~al. 2020.
\newblock {XTREME: A Massively Multilingual Multi-task Benchmark for Evaluating Cross-lingual Generalization}.
\newblock In \emph{Proceedings of ICML}.

\bibitem[{Hu et~al.(2024)Hu, Tu, Han, He, Cui, Long, Zheng, Fang, Huang, Zhao, Zhang, Thai, Zhang, Wang, Yao, Zhao, Zhou, Cai, Zhai, Ding, Jia, Zeng, Li, Liu, and Sun}]{hu2024minicpmunveilingpotentialsmall}
Shengding Hu, Yuge Tu, Xu~Han, Chaoqun He, Ganqu Cui, Xiang Long, Zhi Zheng, Yewei Fang, Yuxiang Huang, Weilin Zhao, Xinrong Zhang, Zheng~Leng Thai, Kaihuo Zhang, Chongyi Wang, Yuan Yao, Chenyang Zhao, Jie Zhou, Jie Cai, Zhongwu Zhai, Ning Ding, Chao Jia, Guoyang Zeng, Dahai Li, Zhiyuan Liu, and Maosong Sun. 2024.
\newblock Minicpm: Unveiling the potential of small language models with scalable training strategies.

\bibitem[{Joshi et~al.(2020)Joshi, Santy, Budhiraja, Bali, and Choudhury}]{joshi2020state}
Pratik Joshi, Sebastin Santy, Amar Budhiraja, Kalika Bali, and Monojit Choudhury. 2020.
\newblock {The State and Fate of Linguistic Diversity and Inclusion in the NLP World}.
\newblock In \emph{Proceedings of ACL}.

\bibitem[{Khashabi et~al.(2018)Khashabi, Chaturvedi, Roth, Upadhyay, and Roth}]{khashabi2018looking}
Daniel Khashabi, Snigdha Chaturvedi, Michael Roth, Shyam Upadhyay, and Dan Roth. 2018.
\newblock {Looking Beyond the Surface: A Challenge Set for Reading Comprehension over Multiple Sentences}.
\newblock In \emph{Proceedings of NAACL-HLT}.

\bibitem[{Lavergne et~al.(2014)Lavergne, Adda, Adda-Decker, and Lamel}]{lavergne-etal-2014-automatic}
Thomas Lavergne, Gilles Adda, Martine Adda-Decker, and Lori Lamel. 2014.
\newblock {Automatic language identity tagging on word and sentence-level in multilingual text sources: a case-study on {L}uxembourgish}.
\newblock In \emph{Proceedings of {LREC}}.

\bibitem[{Levesque et~al.(2012)Levesque, Davis, and Morgenstern}]{levesque2012winograd}
Hector Levesque, Ernest Davis, and Leora Morgenstern. 2012.
\newblock {The Winograd Schema Challenge}.
\newblock In \emph{Proceedings of KR}.

\bibitem[{Liang et~al.(2020)Liang, Gong, Bian, Jiang, Xie, Lin, Feng, Xu, Wang, Chen et~al.}]{liang2020xglue}
Yaobo Liang, Yeyun Gong, Weizhen Bian, Nan Jiang, Guoqing Xie, Ruize Lin, Jiuhai Feng, Ruochen Xu, Wenjie Wang, Zhifang Chen, et~al. 2020.
\newblock {XGLUE: A New Benchmark Dataset for Cross-lingual Pre-training, Understanding and Generation}.
\newblock In \emph{Proceedings of EMNLP}.

\bibitem[{Liu et~al.(2023)Liu, Yuan, Fu, Jiang, Hayashi, and Neubig}]{liu-etal-2023-pretrain}
Pengfei Liu, Weizhe Yuan, Jinlan Fu, Zhengbao Jiang, Hiroaki Hayashi, and Graham Neubig. 2023.
\newblock {Pre-train, Prompt, and Predict: A Systematic Survey of Prompting Methods in Natural Language Processing}.
\newblock \emph{ACM Computing Surveys}, 55(9).

\bibitem[{Liu et~al.(2019)Liu, Ott, Goyal, Du, Joshi, Chen, Levy, Lewis, Zettlemoyer, and Stoyanov}]{liu2019roberta}
Yinhan Liu, Myle Ott, Naman Goyal, Jingfei Du, Mandar Joshi, Danqi Chen, Omer Levy, Mike Lewis, Luke Zettlemoyer, and Veselin Stoyanov. 2019.
\newblock {RoBERTa: A Robustly Optimized BERT Pretraining Approach}.
\newblock In \emph{arXiv preprint arXiv:1907.11692}.

\bibitem[{Lothritz et~al.(2023)Lothritz, Ezzini, Purschke, Bissyande, Klein, Olariu, Boytsov, Lefebvre, and Goujon}]{lothritz2023comparing}
Cedric Lothritz, Saad Ezzini, Christoph Purschke, Tegawend{\'e} Fran{\c{c}}ois D~Assise Bissyande, Jacques Klein, Isabella Olariu, Andrey Boytsov, Clement Lefebvre, and Anne Goujon. 2023.
\newblock {Comparing Pre-Training Schemes for Luxembourgish BERT Models}.
\newblock In \emph{Proceedings of KONVENS}.

\bibitem[{Lothritz et~al.(2022)Lothritz, Lebichot, Allix, Veiber, Bissyande, Klein, Boytsov, Lefebvre, and Goujon}]{lothritz-etal-2022-luxembert}
Cedric Lothritz, Bertrand Lebichot, Kevin Allix, Lisa Veiber, Tegawende Bissyande, Jacques Klein, Andrey Boytsov, Cl{\'e}ment Lefebvre, and Anne Goujon. 2022.
\newblock {LuxemBERT: Simple and Practical Data Augmentation in Language Model Pre-Training for Luxembourgish}.
\newblock In \emph{{Proceedings of LREC}}.

\bibitem[{Lutgen et~al.(2025)Lutgen, Plum, Purschke, and Plank}]{lutgen2025}
Anne-Marie Lutgen, Alistair Plum, Christoph Purschke, and Barbara Plank. 2025.
\newblock {Neural Text Normalization for {L}uxembourgish Using Real-Life Variation Data}.
\newblock In \emph{Proceedings of VarDial}.

\bibitem[{Park et~al.(2021{\natexlab{a}})Park, Zhang, Haley, Steimel, Liu, and Schwartz}]{10.1162/tacl_a_00365}
Hyunji~Hayley Park, Katherine~J. Zhang, Coleman Haley, Kenneth Steimel, Han Liu, and Lane Schwartz. 2021{\natexlab{a}}.
\newblock {Morphology Matters: A Multilingual Language Modeling Analysis}.
\newblock \emph{TACL}.

\bibitem[{Park et~al.(2021{\natexlab{b}})Park, Shin, Lee, Lee, Lee, Lee, Kim, and Kim}]{park2021klue}
Sungjoon Park, Joongbo Shin, Yekyung Lee, Jaehyung Lee, Kichang Lee, Kyunghyun Lee, Sang-Woo Kim, and Heuiseok Kim. 2021{\natexlab{b}}.
\newblock {KLUE: Korean Language Understanding Evaluation}.
\newblock In \emph{Proceedings of NAACL-HLT}.

\bibitem[{Philippy et~al.(2024)Philippy, Haddadan, and Guo}]{philippy2024}
Fred Philippy, Shohreh Haddadan, and Siwen Guo. 2024.
\newblock {Forget {NLI}, Use a Dictionary: Zero-Shot Topic Classification for Low-Resource Languages with Application to {L}uxembourgish}.
\newblock In \emph{Proceedings of SIGUL}.

\bibitem[{Plum et~al.(2026)Plum, Bernardy, and Ranasinghe}]{plum2026ner}
Alistair Plum, Laura Bernardy, and Tharindu Ranasinghe. 2026.
\newblock {Do LLMs Judge Distantly Supervised Named Entity Labels Well? Constructing the JudgeWEL Dataset}.
\newblock In \emph{Proceedings of LREC}.

\bibitem[{Plum et~al.(2024)Plum, D{\"o}hmer, Milano, Lutgen, and Purschke}]{plum2024}
Alistair Plum, Caroline D{\"o}hmer, Emilia Milano, Anne-Marie Lutgen, and Christoph Purschke. 2024.
\newblock {{L}ux{B}ank: The First {U}niversal {D}ependency Treebank for {L}uxembourgish}.
\newblock In \emph{Proceedings of TLT}.

\bibitem[{Plum et~al.(2025)Plum, Ranasinghe, and Purschke}]{plum2025}
Alistair Plum, Tharindu Ranasinghe, and Christoph Purschke. 2025.
\newblock {Text Generation Models for {L}uxembourgish with Limited Data: A Balanced Multilingual Strategy}.
\newblock In \emph{Proceedings of VarDial}.

\bibitem[{Purschke(2020)}]{purschke2020attitudes}
Christoph Purschke. 2020.
\newblock {Attitudes Toward Multilingualism in Luxembourg. A Comparative Analysis of Online News Comments and Crowdsourced Questionnaire Data}.
\newblock \emph{Frontiers in AI}, 3.

\bibitem[{Ranasinghe et~al.(2025)Ranasinghe, Hettiarachchi, Pathirana, Premasiri, Uyangodage, Nanomi~Arachchige, Plum, Rayson, and Mitkov}]{ranasinghe-etal-2025-sinhala}
Tharindu Ranasinghe, Hansi Hettiarachchi, Nadeesha Chathurangi Naradde~Vidana Pathirana, Damith Premasiri, Lasitha Uyangodage, Isuri Nanomi~Arachchige, Alistair Plum, Paul Rayson, and Ruslan Mitkov. 2025.
\newblock {{S}inhala Encoder-only Language Models and Evaluation}.
\newblock In \emph{Proceedings of ACL}.

\bibitem[{Ranasinghe et~al.(2023)Ranasinghe, Plum, Purschke, and Zampieri}]{ranasinghe-etal-2023-publish}
Tharindu Ranasinghe, Alistair Plum, Christoph Purschke, and Marcos Zampieri. 2023.
\newblock {Publish or Hold? {{Automatic}} Comment Moderation in {{Luxembourgish}} News Articles}.
\newblock In \emph{Proceedings of RANLP}.

\bibitem[{Roemmele et~al.(2011)Roemmele, Bejan, and Gordon}]{roemmele2011choice}
Melissa Roemmele, Cosmin~A. Bejan, and Andrew~S. Gordon. 2011.
\newblock {Choice of Plausible Alternatives: An Evaluation of Commonsense Causal Reasoning}.
\newblock In \emph{AAAI Spring Symposium on Logical Formalizations of Commonsense Reasoning}.

\bibitem[{Ruder et~al.(2021)Ruder, Constant, Botha, Siddhant, Firat, Fu, Liu, Hu, Garrette, Neubig, and Johnson}]{ruder2021xtremer}
Sebastian Ruder, Noah Constant, Jan Botha, Aditya Siddhant, Orhan Firat, Jinlan Fu, Pengfei Liu, Junjie Hu, Dan Garrette, Graham Neubig, and Melvin Johnson. 2021.
\newblock {{XTREME}-{R}: Towards More Challenging and Nuanced Multilingual Evaluation}.
\newblock In \emph{Proceedings of EMNLP}.

\bibitem[{Shavrina et~al.(2020)Shavrina, Shevelev, Fenogenova, Nikishina et~al.}]{shavrina-etal-2020-russiansuperglue}
Tatiana Shavrina, Denis Shevelev, Alena Fenogenova, Irina Nikishina, et~al. 2020.
\newblock {RussianSuperGLUE: A Russian Language Understanding Evaluation Benchmark}.
\newblock In \emph{Proceedings of EMNLP}.

\bibitem[{Sirajzade et~al.(2020)Sirajzade, Gierschek, and Schommer}]{sirajzade-etal-2020-sentiment}
Joshgun Sirajzade, Daniela Gierschek, and Christoph Schommer. 2020.
\newblock {An Annotation Framework for Luxembourgish Sentiment Analysis}.
\newblock In \emph{Proceedings of SLTU-CCUR}.

\bibitem[{Snoeren et~al.(2010)Snoeren, Adda-Decker, and Adda}]{snoeren-etal-2010-study}
Natalie~D. Snoeren, Martine Adda-Decker, and Gilles Adda. 2010.
\newblock {The Study of Writing Variants in an Under-resourced Language: Some Evidence from Mobile N-Deletion in {L}uxembourgish}.
\newblock In \emph{Proceedings of {LREC}}.

\bibitem[{Socher et~al.(2013)Socher, Perelygin, Wu, Chuang, Manning, Ng, and Potts}]{socher2013recursive}
Richard Socher, Alex Perelygin, Jean Wu, Jason Chuang, Christopher~D. Manning, Andrew~Y. Ng, and Christopher Potts. 2013.
\newblock {Recursive deep models for semantic compositionality over a sentiment treebank}.
\newblock In \emph{Proceedings of EMNLP}.

\bibitem[{Tay et~al.(2022)Tay, Dehghani, Bahri, and Metzler}]{tay2020efficient}
Yi~Tay, Mostafa Dehghani, Dara Bahri, and Donald Metzler. 2022.
\newblock {Efficient Transformers: A Survey}.
\newblock \emph{ACM Computing Surveys}, 55(6).

\bibitem[{van~der Goot et~al.(2021)van~der Goot, Sharaf, Imankulova, {\"U}st{\"u}n, Stepanovi{\'c}, Ramponi, Khairunnisa, Komachi, and Plank}]{van-der-goot-etal-2021-masked}
Rob van~der Goot, Ibrahim Sharaf, Aizhan Imankulova, Ahmet {\"U}st{\"u}n, Marija Stepanovi{\'c}, Alan Ramponi, Siti~Oryza Khairunnisa, Mamoru Komachi, and Barbara Plank. 2021.
\newblock {From Masked Language Modeling to Translation: Non-{E}nglish Auxiliary Tasks Improve Zero-shot Spoken Language Understanding}.
\newblock In \emph{Proceedings of NAACL-HLT}.

\bibitem[{Vaswani et~al.(2017)Vaswani, Shazeer, Parmar, Uszkoreit, Jones, Gomez, Kaiser, and Polosukhin}]{vaswani2017attention}
Ashish Vaswani, Noam Shazeer, Niki Parmar, Jakob Uszkoreit, Llion Jones, Aidan~N Gomez, {\L}ukasz Kaiser, and Illia Polosukhin. 2017.
\newblock {Attention is all you need}.
\newblock In \emph{Proceedings of NeurIPS}.

\bibitem[{Wang et~al.(2019{\natexlab{a}})Wang, Pruksachatkun, Nangia, Singh, Michael, Hill, Levy, and Bowman}]{wang19superglue}
Alex Wang, Yada Pruksachatkun, Nikita Nangia, Amanpreet Singh, Julian Michael, Felix Hill, Omer Levy, and Samuel~R. Bowman. 2019{\natexlab{a}}.
\newblock {SuperGLUE: A Stickier Benchmark for General-Purpose Language Understanding Systems}.
\newblock In \emph{Proceedings of NeurIPS}.

\bibitem[{Wang et~al.(2019{\natexlab{b}})Wang, Singh, Michael, Hill, Levy, and Bowman}]{wang-etal-2018-glue}
Alex Wang, Amanpreet Singh, Julian Michael, Felix Hill, Omer Levy, and Samuel~R. Bowman. 2019{\natexlab{b}}.
\newblock {GLUE: A Multi-Task Benchmark and Analysis Platform for Natural Language Understanding}.
\newblock In \emph{Proceedings of ICLR}.

\bibitem[{Warner et~al.(2025)Warner, Chaffin, Clavi{\'e}, Weller, Hallstr{\"o}m, Taghadouini, Gallagher, Biswas, Ladhak, Aarsen, Adams, Howard, and Poli}]{warner2024smarterbetterfasterlonger}
Benjamin Warner, Antoine Chaffin, Benjamin Clavi{\'e}, Orion Weller, Oskar Hallstr{\"o}m, Said Taghadouini, Alexis Gallagher, Raja Biswas, Faisal Ladhak, Tom Aarsen, Griffin~Thomas Adams, Jeremy Howard, and Iacopo Poli. 2025.
\newblock Smarter, better, faster, longer: A modern bidirectional encoder for fast, memory efficient, and long context finetuning and inference.
\newblock In \emph{Proceedings of ACL}.

\bibitem[{Wei et~al.(2022)Wei, Wang, Schuurmans, Bosma, Chi, Le, and Zhou}]{wei-etal-2022-chain}
Jason Wei, Xuezhi Wang, Dale Schuurmans, Maarten Bosma, Ed~Chi, Quoc Le, and Denny Zhou. 2022.
\newblock {Chain-of-Thought Prompting Elicits Reasoning in Large Language Models}.
\newblock In \emph{Proceedings of NeurIPS}.

\bibitem[{Weller et~al.(2026)Weller, Ricci, Marone, Chaffin, Lawrie, and Durme}]{weller2025seqvsseqopen}
Orion Weller, Kathryn Ricci, Marc Marone, Antoine Chaffin, Dawn Lawrie, and Benjamin~Van Durme. 2026.
\newblock {Seq vs Seq: An Open Suite of Paired Encoders and Decoders}.
\newblock In \emph{Proceedings of ICLR}.

\bibitem[{Williams et~al.(2018)Williams, Nangia, and Bowman}]{williams2018broad}
Adina Williams, Nikita Nangia, and Samuel~R. Bowman. 2018.
\newblock {A Broad-Coverage Challenge Corpus for Sentence Understanding through Inference}.
\newblock In \emph{Proceedings of NAACL-HLT}.

\bibitem[{Wortsman et~al.(2023)Wortsman, Dettmers, Zettlemoyer, Morcos, Farhadi, and Schmidt}]{wortsman2023stablelowprecisiontraininglargescale}
Mitchell Wortsman, Tim Dettmers, Luke Zettlemoyer, Ari Morcos, Ali Farhadi, and Ludwig Schmidt. 2023.
\newblock {Stable and low-precision training for large-scale vision-language models}.
\newblock In \emph{Proccedings of NeurIPS}.

\bibitem[{Wu and Dredze(2019)}]{wu-dredze-2019-beto}
Shijie Wu and Mark Dredze. 2019.
\newblock {Beto, Bentz, Becas: The Surprising Cross-Lingual Effectiveness of BERT}.
\newblock In \emph{Proceedings of EMNLP-IJCNLP}.

\bibitem[{Yang et~al.(2019)Yang, Dai, Yang, Carbonell, Salakhutdinov, and Le}]{yang2019xlnet}
Zhilin Yang, Zihang Dai, Yiming Yang, Jaime Carbonell, Ruslan Salakhutdinov, and Quoc~V. Le. 2019.
\newblock {XLNet: Generalized Autoregressive Pretraining for Language Understanding}.
\newblock In \emph{Proceedings of NeurIPS}.

\bibitem[{Zhai et~al.(2022)Zhai, Kolesnikov, Houlsby, and Beyer}]{zhai2022scalingvisiontransformers}
Xiaohua Zhai, Alexander Kolesnikov, Neil Houlsby, and Lucas Beyer. 2022.
\newblock {Scaling Vision Transformers}.
\newblock In \emph{Proceedings of CVPR}.

\bibitem[{Zhang et~al.(2023)Zhang, Li, Hauer, Shi, and Kondrak}]{zhang2023donttrustchatgptquestion}
Xiang Zhang, Senyu Li, Bradley Hauer, Ning Shi, and Grzegorz Kondrak. 2023.
\newblock {Don't Trust ChatGPT when Your Question is not in English: A Study of Multilingual Abilities and Types of LLMs}.
\newblock In \emph{Proceedings of EMNLP}.

\end{thebibliography}

\section{Appendix}\label{sec:appendix}
\crefalias{section}{appendix}
\subsection{\texttt{ltzGLUE} Task Examples}

For demonstration purposes, we present an example for each task in \texttt{ltzGLUE} in Table \ref{tab:task_examples}. The examples are intended to illustrate the task formulations and typical model inputs and outputs.

\begin{table*}[t]
\centering
\small
\begin{tabular}{p{0.5cm} p{12.6cm}}
\toprule
\textbf{Task} & \textbf{Content} \\
\midrule

\hc
& \textbf{Input:} \textit{Headline:} Paschtouer krut 2.500 Euro vun Onéierlechen ofgeknäppt
  \textit{(Priest robbed of 2500€ by criminals)}\newline
  \textit{Article:} Déi lescht Wochen hätt een ëmmer méi dacks Kollekte gemellt kritt, déi awer net vun Handicap International an Optrag gi goufen…
  \textit{(In the past few weeks, an increasing number of charity collections were reported, which were not commissioned by Handicap International)} \\
& \textbf{Output:} correct \\
\midrule

\sa
& \textbf{Input:} Et war den Houfert vun e puer Generatioune Lëtzebuerger.
  \textit{(It was the pride of a couple of Luxembourgish generations)} \\
& \textbf{Output:} positive \\
\midrule

\labs
& \textbf{Input:} Dat Bild do ass eng plomper Fälschung!
  \textit{(That painting is an amateurish forgery!)} \\
& \textbf{Output:} incorrect \\
\midrule

\lams
& \textbf{Input:} Ech schonn dräimol Usbekistan.
  \textit{I (have) been (to) Uzbekistan three times.} \\
& \textbf{Output:} syntax \\
\midrule

\ner
& \textbf{Input:} De Mark Cavendish konnt de Massesprint op der Arrivée fir sech entscheeden.
  \textit{(Mark Cavendish was able to win the mass sprint on the finish line.)} \\
& \textbf{Output:} "O", "B-PER", "I-PER", "O", "O", "O", "O", "O", "O", "O", "O", "O", "O" \\
\midrule

\tc
& \textbf{Input:} Kënnt Dir Iech nach un de Meister Ede a säi Pumuckl erënneren?…
  \textit{(Do you remember Master Ede and his Pumuckl?)} \\
& \textbf{Output:} culture \\
\midrule

\sid
& \textbf{Input:} Reent et nächst Woch?
  \textit{(Will it rain next week?)} \\
& \textbf{Output:} weather/find \\
\midrule

\rte
& \textbf{Input:} IBM huet Geschäftsgeheimnisser geklaut, fir zwee vu senge Programmer ze kopéieren: File-AID, en Dateiemanager, an Abend-AID, e Programm, deen de Benotzer hëlleft, d’Quell vu Feeler ze fannen.
  \textit{(IBM has stolen confidential business files to copy two programs: File-AID, a data manager, and Abend-AID, a program to detect sources of mistakes.)}\newline
  Geschäftsgeheimnisser goufen geklaut.
  \textit{(Business secrets were stolen.)} \\
& \textbf{Output:} true \\
\bottomrule

\end{tabular}
\caption{\textbf{Input--output examples for each task.}
LTZ inputs are shown with English translations for clarity.}
\label{tab:task_examples}
\end{table*}

\subsection{\modelname Training Details}
\label{app:arch}
\begin{table}[h]
\centering
\small
\begin{tabular}{@{}p{0.5\columnwidth}r@{}}
\toprule
\textbf{Parameter} & \textbf{Value} \\
\midrule
Vocabulary Size & 50,368 \\
Max Sequence Length & 1024 \\
Tokenizer Arch. & GPTNeoXTokenizerFast \\
Attention Layer & RoPE \\
Attention Dropout & 0.0 \\
Attention Output Bias & false \\
Attention Output Dropout & 0.1 \\
Attention QKV Bias & false \\
Transformer Layer & prenorm \\
Embedding Dropout & 0.0 \\
Embedding Norm & true \\
Final Norm & true \\
Skip First PreNorm & true \\
Embedding Layer & sans\_pos \\
MLP Dropout & 0.0 \\
MLP Input Bias & false \\
MLP Layer Type & GLU \\
MLP Output Bias & false \\
MLM Probability & 0.3 \\
Normalization & LayerNorm \\
Norm Epsilon & 1e-5 \\
Norm Bias & false \\
Hidden Activation & GELU \\
Head Pred Activation & GELU \\
Activation Function & GELU \\
Padding & unpadded \\
Rotary Embedding Base & 10,000.0 \\
Rotary Embedding Interleaved & false \\
Allow Embedding Resizing & true \\
Sliding Window & 128 \\
Global Attn. every N Layers & 3 \\
Unpad Embeddings & true \\
Masked Prediction & true \\
\bottomrule
\end{tabular}
\caption{Common pre-training configuration parameters across both \modelname models (mini and base).}
\label{tab:pre-training_config}
\end{table}
\paragraph{Model Architecture}
We follow the Ettin recipe \cite{weller2025seqvsseqopen}, based on ModernBERT \cite{warner2024smarterbetterfasterlonger}, for training hyperpameters and model architecture. We train two sizes of \modelname models, mini and base, with 68M and 110M non-embedding parameters, respectively. Common pre-training configuration parameters for both sizes can be found in \cref{tab:pre-training_config}. \modelname-mini has 19 hidden layers, a hidden size of 512, an intermediate size of 768, and 8 attention heads, whereas \modelname-base has 22 hidden layers, a hidden size of 768, an intermediate size of 1152, and 12 attention heads.

Both models share a GPTNeoXTokenizerFast tokenizer \cite{black-etal-2022-gpt} \footnote{\url{https://huggingface.co/docs/transformers/v4.57.3/en/model_doc/gpt_neox}}, a BPE-based tokenizer, which we train on the entire pre-training set, using a minimum frequency of two and a vocabulary size of 50,368.

\paragraph{Training Details}
We use a constant batch size of 1024 packed sequences, where both models have a max sequence length of 1024. We follow ModernBERT \cite{warner2024smarterbetterfasterlonger} and Ettin \cite{weller2025seqvsseqopen} in using the Warmup-Stable-Decay (WSD) scheduler \cite{zhai2022scalingvisiontransformers, hu2024minicpmunveilingpotentialsmall}, though we use a shorter warmup and decay phase of 500 batches each, due to our smaller pre-training dataset size and larger number of epochs (10 vs. one). Again following ModernBERT and Ettin's recipe, we use the StableAdamW optimizer \cite{wortsman2023stablelowprecisiontraininglargescale}, with a peak learning rate of 3e-3 with a weight decay of 3e-4 for \modelname-mini and 8e-4 with a weight decay of 1e-5 for \modelname-base. As our pre-training set is small, we train each model for 10 epochs, following \citet{lothritz-etal-2022-luxembert}.

\paragraph{Computational Resources}
We use a 20GB MIG partition of an NVIDIA A100-SXM4-80GB to pretrain each model, taking 47 hours for \modelname-mini and 76 hours for \modelname-base. However, we note that compute times were negatively impacted by concurrent jobs on the server cluster with suboptimal CPU thread management.

\paragraph{Pre-training Data Breakdown}\label{app:pre-training-counts}
We show pre-training data token counts per source in \cref{tab:pre-training-counts}, where sources (described in \cref{sec:model}) are: RTL news articles (News), RTL transcribed radio interviews (Radio), RTL user comments (Comments), transcribed podcasts (Podcasts), transcribed political speeches and debates from the Chambre des Députés (Chamber), 1M sentences from the web crawl of the Leipzig Collection (Web), text from Luxembourgish chat rooms (Webchat), a Wikipedia crawl (Wikipedia), and examples from the Luxembourgish Online Dictionary (LOD).

\begin{table}[]
\centering
\small
\begin{tabular}{lr}
\toprule
\textbf{Source} & \multicolumn{1}{l}{\textbf{Tokens (M)}} \\
\midrule
Wikipedia & 11.2 \\
LOD & 0.7 \\
RTL Radio & 24.9 \\
RTL News & 51.6 \\
RTL Comments & 77.7 \\
Chamber & 23.5 \\
Web & 22.9 \\
Webchat & 20.8 \\
\midrule
\textbf{Total} & 233.4 \\
\bottomrule
\end{tabular}
\caption{Token counts (M) per source for pretraining data of \modelname.}
\label{tab:pre-training-counts}
\end{table}

\subsection{Hyperparameter Sweeps}\label{app:sweeps}
\begin{table}[]
\centering
\small
\begin{tabular}{@{}ll@{}}
\toprule
\textbf{Parameter} & \textbf{Values} \\
\midrule
Learning Rate & \{1e-5, 3e-5, 5e-5, 8e-5\} \\
Batch Size & \{8, 16\} \\
Epochs & \{2, 5\} \\
Weight Decay & \{0.0, 0.01\} \\
Warmup Ratio & \{0.0, 0.1\} \\
\bottomrule
\end{tabular}
\caption{Hyperparameter sweep ranges used for all task and model combinations.}
\label{tab:sweep-ranges}
\end{table}
Though we do not aim to optimise performance in our evaluation, we conduct basic hyperparameter sweeps for each model and task combination in order to provide a fairer comparison across models. We use Weights \& Biases version 0.23.1. to conduct the sweeps. For each model and task combination, we select the best hyperparameters based on the validation set, and use those parameters to fine-tune two additional models with differing seeds, resulting in three runs. In order to reduce the computational demand of the sweeps, we use Bayesian search with early stopping after three iterations, and cap each sweep at 30 runs, for 1,440 total runs across all models and tasks (and an additional 96 to finetune the two additional seeds). For each sweep we use the same hyperparameter ranges, shown in \cref{tab:sweep-ranges}. Best values for each sweep are shown in \cref{tab:sweep-results}. However, we note again that these ranges were kept simple to keep sweeps computationally feasible, thus, these values should not be seen as optimal hyperparameters.

\begin{table*}[h]
\small
\setlength{\tabcolsep}{3pt}
\centering
\begin{tabular}{@{}llrrrrr@{}}
\toprule
\textbf{Task} & \textbf{Model} & \textbf{Learning Rate} & \textbf{Batch Size} & \textbf{Epochs} & \textbf{Weight Decay} & \textbf{Warmup Ratio} \\
\midrule
\hc & \luxembert & 5e-5 & 16 & 2 & 0 & 0 \\
\hc & \mbert & 1e-5 & 16 & 2 & 0.01 & 0.1 \\
\hc & \modelname-base & 3e-5 & 8 & 5 & 0 & 0.1 \\
\hc & \modelname-mini & 8e-5 & 8 & 5 & 0.01 & 0.1 \\
\hc & \mmbert & 1e-5 & 8 & 5 & 0 & 0 \\
\hc & \roberta & 1e-5 & 8 & 5 & 0 & 0.1 \\
\midrule
\sa & \luxembert & 8e-5 & 16 & 2 & 0 & 0 \\
\sa & \mbert & 1e-5 & 16 & 5 & 0 & 0.1 \\
\sa & \modelname-base & 8e-5 & 16 & 5 & 0 & 0 \\
\sa & \modelname-mini & 8e-5 & 8 & 5 & 0 & 0.1 \\
\sa & \mmbert & 1e-5 & 16 & 2 & 0.01 & 0.1 \\
\sa & \roberta & 1e-5 & 8 & 5 & 0.01 & 0 \\
\midrule
\labs & \luxembert & 8e-5 & 16 & 2 & 0 & 0 \\
\labs & \mbert & 1e-5 & 16 & 5 & 0 & 0.1 \\
\labs & \modelname-base & 8e-5 & 8 & 5 & 0.01 & 0.1 \\
\labs & \modelname-mini & 8e-5 & 16 & 5 & 0 & 0 \\
\labs & \mmbert & 3e-5 & 16 & 5 & 0 & 0.1 \\
\labs & \roberta & 1e-5 & 16 & 5 & 0.01 & 0 \\
\midrule
\lams & \luxembert & 5e-5 & 16 & 2 & 0.01 & 0.1 \\
\lams & \mbert & 3e-5 & 16 & 5 & 0.01 & 0.1 \\
\lams & \modelname-base & 8e-5 & 8 & 5 & 0.01 & 0.1 \\
\lams & \modelname-mini & 8e-5 & 8 & 5 & 0.01 & 0.1 \\
\lams & \mmbert & 3e-5 & 16 & 5 & 0 & 0.1 \\
\lams & \roberta & 1e-5 & 8 & 5 & 0.01 & 0.1 \\
\midrule
\ner & \luxembert & 3e-5 & 16 & 5 & 0 & 0.1 \\
\ner & \mbert & 3e-5 & 8 & 5 & 0 & 0 \\
\ner & \modelname-base & 8e-5 & 8 & 5 & 0.01 & 0.1 \\
\ner & \modelname-mini & 8e-5 & 8 & 5 & 0.01 & 0.1 \\
\ner & \mmbert & 5e-5 & 16 & 5 & 0.01 & 0.1 \\
\ner & \roberta & 5e-5 & 16 & 5 & 0 & 0.1 \\
\midrule
\sid & \luxembert & 3e-5 & 16 & 2 & 0.01 & 0 \\
\sid & \mbert & 3e-5 & 8 & 5 & 0 & 0 \\
\sid & \modelname-base & 8e-5 & 8 & 5 & 0 & 0 \\
\sid & \modelname-mini & 8e-5 & 8 & 5 & 0 & 0 \\
\sid & \mmbert & 8e-5 & 16 & 2 & 0 & 0 \\
\sid & \roberta & 3e-5 & 16 & 5 & 0 & 0 \\
\midrule
\tc & \luxembert & 3e-5 & 8 & 5 & 0.01 & 0 \\
\tc & \mbert & 1e-5 & 8 & 5 & 0.01 & 0.1 \\
\tc & \modelname-base & 8e-5 & 16 & 5 & 0.01 & 0.1 \\
\tc & \modelname-mini & 8e-5 & 8 & 5 & 0.01 & 0 \\
\tc & \mmbert & 3e-5 & 8 & 5 & 0.01 & 0.1 \\
\tc & \roberta & 1e-5 & 16 & 5 & 0.01 & 0 \\
\midrule
\rte & \luxembert & 8e-5 & 16 & 2 & 0 & 0 \\
\rte & \mbert & 1e-5 & 16 & 5 & 0.01 & 0.1 \\
\rte & \modelname-base & 8e-5 & 8 & 2 & 0 & 0.1 \\
\rte & \modelname-mini & 5e-5 & 16 & 5 & 0.01 & 0 \\
\rte & \mmbert & 3e-5 & 8 & 5 & 0 & 0.1 \\
\rte & \roberta & 1e-5 & 16 & 2 & 0.01 & 0.1 \\
\bottomrule
\end{tabular}
\caption{\textbf{Best hyperparameters per model for each task.}}
\label{tab:sweep-results}
\end{table*}

\paragraph{Computational Resources}
We use several 20GB MIG partitions of NVIDIA A100-SXM4-80GB GPUs to conduct the sweeps. Depending on model and task dataset size, multiple runs were conducted in parallel on each partition, totalling 59 days of compute, which includes fine-tuning the additional seeds, as well as evaluation on the validation and test sets.

\subsection{Prompt to Improve Quality of RTE Task}
\label{app:improve_rte}
\begin{verbatim}
You are an expert for the Luxembourgish 
language. I am giving you a sentence in 
Luxembourgish. You have to judge its 
quality and improve it while keeping
the meaning intact. As output, write only 
the improved sentence or the original 
sentence if it is of very high quality.
\end{verbatim}

\subsection{Prompt to Judge the Quality of Improved RTE Dataset}
\label{app:judge_rte}
\begin{verbatim}
You are an expert for the Luxembourgish 
language. I am giving you two texts in 
Luxembourgish. You have to judge their 
quality. As output, simply write 'low', 
'medium' or 'high' depending on the 
quality of both sentences, nothing else.    
\end{verbatim}

\subsection{Prompt to Verify the Labels of Improved RTE Dataset}
\label{app:verify_rte_labels}
\begin{verbatim}
You are an expert for the Luxembourgish 
language. I am giving you two texts 
TEXT1 and TEXT2 in Luxembourgish as well 
as a LABEL where 1 means that TEXT1 
logically entails TEXT2 while 0 means the 
opposite. You have to check if the labels 
are correct. As output, simply write 'true' 
if the label is the correct one or 'false' 
if the label is incorrect.
\end{verbatim}

\subsection{Main prompt for zero-shot testing of LLMs}
\label{app:zeroshot_llms}
\begin{verbatim}
You are a classification and text-processing 
model specialized in NLP tasks for 
Luxembourgish (lb).
Follow ALL rules strictly:
1. Respond ONLY in valid JSON.
2. Do NOT add explanations, comments or text 
outside of JSON.
3. Use field: "output": <model_answer>.
4. Use field: "task": "<task_name>".
5. Use field: "input": "<input example text>".
6. Predict only the requested outputs and 
label(s) in the given formats.
7. If determined labels are 0 and 1 then 0 
is used for False, 1 is used for True.
Here is the NLP task definition:
TASK: {task_name}
DESCRIPTION: {task_description}
\end{verbatim}

\subsection{Task descriptions for zero-shot testing of LLMs}
\label{app:zeroshot_llms_tasks}
\begin{verbatim}
headline_classification:
Decide if the given title/headline fits the 
text. 
Output True or False.

sentiment_analysis: 
Classify sentiment of the text. 
Allowed labels: positive, neutral, negative.

linguistic_acceptability_binary: 
Decide whether the sentence is linguis-
tically acceptable in Luxembourgish.
Output: 0 or 1.
       
linguistic_acceptability_multilabel: 
Detect if the sentence is correct or if 
some element is wrong. 
If the sentence is correct, 
Output: correct. 
If it is not, Output the label referencing 
the wrong element: 
syntax, verb, ortho or adj.

ner: 
Perform Named Entity Recognition
on the given sequence of sentence 
tokens.
Output tags as lists of ner_tags.
Allowed Tags: O, B-LOC, I-LOC,
B-PER,I-PER, B-DATE, I-DATE,B-ORG, 
I-ORG, B-MISC, I-MISC.

topic_classification: 
Classify topic of the document 
by title and text. 
Allowed category_names: sports, 
animals, business, culture, technology.

slot_intent_detection: 
Detect the intent for
the text given.
Allowed intents: 
reminder/show_reminders, 
weather/find\, 
reminder/set_reminder, 
reminder/cancel_reminder,
alarm/snooze_alarm,
alarm/show_alarms, 
alarm/set_alarm, 
nalarm/cancel_alarm, 
nalarm/time_left_on_alarm.

recognizing_textual_entailment: 
Determine if the information in the second 
sentence is entailed in the first one. 
Output: 0 or 1.
\end{verbatim}

\subsection{Full Results}\label{app:full}
We show full results (validation and test set performance) for each model and task for \hc, \sa, \lab, and \lam in \cref{tab:full_results} and for \ner, \tc, \sid, and \rte in \cref{tab:full_results_ner_tc_sid_rte}.
\begin{table}[h]
\centering
\small
\begin{tabular}{@{}llrrrr@{}}
\toprule
\textbf{Task} & \textbf{Model} & \textbf{Dev \Fone} & \textbf{Test \Fone}\\
\midrule
\hc & \luxembert & \cellcolor[HTML]{E2F4EB}\fonesd{66.18}{0.00} & \cellcolor[HTML]{E5F5ED}\fonesd{66.37}{0.00}\\
\hc & \mbert & \cellcolor[HTML]{8DD1B0}\fonesd{77.50}{10.05} & \cellcolor[HTML]{90D2B2}\fonesd{77.91}{10.26}\\
\hc & \modelname-base & \cellcolor[HTML]{FFFFFF}\fonesd{62.29}{4.91} & \cellcolor[HTML]{FFFFFF}\fonesd{62.81}{4.98}\\
\hc & \modelname-mini & \cellcolor[HTML]{B1E0C9}\fonesd{72.75}{1.15} & \cellcolor[HTML]{B7E2CD}\fonesd{72.69}{.33}\\
\hc & \mmbert & \cellcolor[HTML]{57BB8A}\fonesd{84.56}{2.66} & \cellcolor[HTML]{57BB8A}\fonesd{85.59}{1.61}\\
\hc & \roberta & \cellcolor[HTML]{B7E2CD}\fonesd{71.91}{9.92} & \cellcolor[HTML]{BBE4D0}\fonesd{72.09}{9.90}\\
\midrule
\sa & \luxembert & \cellcolor[HTML]{57BB8A}\fonesd{58.78}{2.27} & \cellcolor[HTML]{57BB8A}\fonesd{58.66}{0.73}\\
\sa & \mbert & \cellcolor[HTML]{D4EEE1}\fonesd{43.99}{10.42} & \cellcolor[HTML]{DBF1E6}\fonesd{41.25}{4.87}\\
\sa & \modelname-base & \cellcolor[HTML]{C3E7D5}\fonesd{46.05}{9.44} & \cellcolor[HTML]{B3E0CA}\fonesd{46.59}{5.88}\\
\sa & \modelname-mini & \cellcolor[HTML]{C0E6D3}\fonesd{46.45}{8.10} & \cellcolor[HTML]{BCE4D0}\fonesd{45.39}{6.79}\\
\sa & \mmbert & \cellcolor[HTML]{70C59C}\fonesd{55.86}{4.18} & \cellcolor[HTML]{7FCCA6}\fonesd{53.37}{4.47}\\
\sa & \roberta & \cellcolor[HTML]{FFFFFF}\fonesd{38.87}{1.29} & \cellcolor[HTML]{FFFFFF}\fonesd{36.40}{0.57}\\
\midrule
\labs & \luxembert & \cellcolor[HTML]{5BBD8D}\fonesd{89.61}{0.69} & \cellcolor[HTML]{5EBE8F}\fonesd{89.17}{0.18}\\
\labs & \mbert & \cellcolor[HTML]{97D5B7}\fonesd{82.74}{0.73} & \cellcolor[HTML]{A2DABE}\fonesd{81.26}{0.56}\\
\labs & \modelname-base & \cellcolor[HTML]{8BD0AE}\fonesd{84.13}{7.57} & \cellcolor[HTML]{91D3B3}\fonesd{83.17}{8.38}\\
\labs & \modelname-mini & \cellcolor[HTML]{57BB8A}\fonesd{90.00}{2.64} & \cellcolor[HTML]{5DBE8E}\fonesd{89.31}{2.89}\\
\labs & \mmbert & \cellcolor[HTML]{5ABD8C}\fonesd{89.71}{ 0.20} & \cellcolor[HTML]{57BB8A}\fonesd{89.97}{0.09}\\
\labs & \roberta & \cellcolor[HTML]{FFFFFF}\fonesd{70.86}{4.47} & \cellcolor[HTML]{FFFFFF}\fonesd{70.}{4.55}\\
\midrule
\lams & \luxembert & \cellcolor[HTML]{6EC59A}\fonesd{88.67}{0.31} & \cellcolor[HTML]{6EC59A}\fonesd{87.96}{0.71}\\
\lams & \mbert & \cellcolor[HTML]{AEDFC7}\fonesd{82.02}{1.14} & \cellcolor[HTML]{AEDFC7}\fonesd{81.20}{1.02}\\
\lams & \modelname-base & \cellcolor[HTML]{CAEADA}\fonesd{79.19}{11.94} & \cellcolor[HTML]{C6E8D8}\fonesd{78.63}{11.55}\\
\lams & \modelname-mini & \cellcolor[HTML]{81CCA8}\fonesd{86.70}{4.59} & \cellcolor[HTML]{7BCAA3}\fonesd{86.62}{4.64}\\
\lams & \mmbert & \cellcolor[HTML]{57BB8A}\fonesd{91.03}{0.83} & \cellcolor[HTML]{57BB8A}\fonesd{90.35}{0.72}\\
\lams & \roberta & \cellcolor[HTML]{FFFFFF}\fonesd{73.58}{3.31} & \cellcolor[HTML]{FFFFFF}\fonesd{72.59}{2.69}\\
\bottomrule
\end{tabular}
\caption{\textbf{Dev and Test \Fone scores for Headline Acceptability (\hc), Sentiment Analysis (\sa) and Linguistic Acceptability (Binary \labs and Multi \lams}. Results are averaged over three runs, with standard deviations as subscripts.}
\label{tab:full_results}
\end{table}
\begin{table}[h]
\centering
\small
\begin{tabular}{@{}llrrrr@{}}
\toprule
\textbf{Task} & \textbf{Model} & \textbf{Dev \Fone} & \textbf{Test \Fone}\\
\midrule
\ner & \luxembert & \cellcolor[HTML]{81CCA7}\fonesd{89.44}{0.35} & \cellcolor[HTML]{6BC398}\fonesd{90.57}{0.09}\\
\ner & \mbert & \cellcolor[HTML]{57BB8A}\fonesd{90.28}{0.44} & \cellcolor[HTML]{5FBE90}\fonesd{90.83}{0.28}\\
\ner & \modelname-base & \cellcolor[HTML]{C2E7D5}\fonesd{88.12}{0.99} & \cellcolor[HTML]{AEDFC7}\fonesd{89.17}{0.70}\\
\ner & \modelname-mini & \cellcolor[HTML]{FFFFFF}\fonesd{86.86}{0.79} & \cellcolor[HTML]{FFFFFF}\fonesd{87.47}{0.37}\\
\ner & \mmbert & \cellcolor[HTML]{64C093}\fonesd{90.03}{0.49} & \cellcolor[HTML]{57BB8A}\fonesd{90.98}{0.23}\\
\ner & \roberta & \cellcolor[HTML]{A9DCC3}\fonesd{88.63}{1.00} & \cellcolor[HTML]{90D2B2}\fonesd{89.80}{0.50}\\
\midrule
\tc & \luxembert & \cellcolor[HTML]{5BBD8D}\fonesd{98.36}{0.26} & \cellcolor[HTML]{72C69D}\fonesd{98.68}{0.17}\\
\tc & \mbert & \cellcolor[HTML]{D9F0E4}\fonesd{97.27}{1.26} & \cellcolor[HTML]{C8E9D9}\fonesd{97.80}{0.67}\\
\tc & \modelname-base & \cellcolor[HTML]{8ED2B1}\fonesd{97.91}{0.40} & \cellcolor[HTML]{57BB8A}\fonesd{98.95}{0.36}\\
\tc & \modelname-mini & \cellcolor[HTML]{83CDA9}\fonesd{98.01}{0.52} & \cellcolor[HTML]{84CDA9}\fonesd{98.50}{0.23}\\
\tc & \mmbert& \cellcolor[HTML]{57BB8A}\fonesd{98.38}{0.29} & \cellcolor[HTML]{5BBD8D}\fonesd{98.92}{0.28}\\
\tc & \roberta & \cellcolor[HTML]{FFFFFF}\fonesd{96.94}{1.33} & \cellcolor[HTML]{FFFFFF}\fonesd{97.24}{0.50}\\
\midrule
\sid & \luxembert & \cellcolor[HTML]{57BB8A}\fonesd{100.00}{0.00} & \cellcolor[HTML]{57BB8A}\fonesd{91.71}{0.11}\\
\sid & \mbert & \cellcolor[HTML]{DCF1E6}\fonesd{81.02}{6.52} & \cellcolor[HTML]{FFFFFF}\fonesd{60.65}{4.71}\\
\sid & \modelname-base & \cellcolor[HTML]{C1E6D4}\fonesd{84.82}{13.43} & \cellcolor[HTML]{BBE4D0}\fonesd{73.32}{11.66}\\
\sid & \modelname-mini & \cellcolor[HTML]{ECF8F2}\fonesd{78.64}{18.57} & \cellcolor[HTML]{96D5B6}\fonesd{80.13}{3.00}\\
\sid & \mmbert & \cellcolor[HTML]{CEEBDD}\fonesd{83.03}{15.22} & \cellcolor[HTML]{A0D9BD}\fonesd{78.26}{7.22}\\
\sid & \roberta & \cellcolor[HTML]{FFFFFF}\fonesd{75.85}{6.98} & \cellcolor[HTML]{F4FBF8}\fonesd{62.75}{8.76}\\
\midrule
\rte & \luxembert & \cellcolor[HTML]{91D3B3}\fonesd{70.31}{0.71} & \cellcolor[HTML]{94D4B5}\fonesd{70.96}{0.26}\\
\rte & \mbert & \cellcolor[HTML]{78C9A1}\fonesd{72.12}{1.37} & \cellcolor[HTML]{72C69D}\fonesd{73.48}{1.84}\\
\rte & \modelname-base & \cellcolor[HTML]{B3E1CA}\fonesd{67.84}{4.34} & \cellcolor[HTML]{BEE5D2}\fonesd{67.88}{3.20}\\
\rte & \modelname-mini & \cellcolor[HTML]{FFFFFF}\fonesd{62.33}{4.05} & \cellcolor[HTML]{FFFFFF}\fonesd{63.10}{3.64}\\
\rte & \mmbert & \cellcolor[HTML]{57BB8A}\fonesd{74.49}{1.81} & \cellcolor[HTML]{57BB8A}\fonesd{75.44}{2.41}\\
\rte & \roberta & \cellcolor[HTML]{85CEAA}\fonesd{71.22}{0.70} & \cellcolor[HTML]{96D5B6}\fonesd{70.82}{0.42}\\
\bottomrule
\end{tabular}
\caption{\textbf{Dev and Test \Fone scores for Named Entity Recognition (\ner), Topic Classification (\tc), Intent Detection (\sid) and Textual Entailment (\rte)}. Results are averaged over three runs, with standard deviations as subscripts.}
\label{tab:full_results_ner_tc_sid_rte}
\end{table}

\end{document}